\documentclass{article} %
\usepackage{colm2024_conference}

\usepackage{microtype}
\usepackage{hyperref}
\usepackage{url}
\usepackage{booktabs}
\definecolor{darkblue}{rgb}{0, 0, 0.5}
\hypersetup{colorlinks=true, citecolor=darkblue, linkcolor=darkblue, urlcolor=darkblue}

\usepackage{colortbl}
\usepackage{tablefootnote}
\usepackage{graphicx}
\usepackage{multirow}
\usepackage{enumitem}
\usepackage{wrapfig}

\newcommand\raven{\raisebox{-2pt}{\includegraphics[width=0.9em]{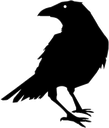}}}

\newcommand{\mask}[1]{\texttt{<extra\_id\_{#1}>}}

\newcommand{\p}[1]{{\textbf{#1}}}

\title{\raven{} \textsc{Raven}: In-Context Learning with Retrieval-Augmented Encoder-Decoder Language Models}

\author{Jie Huang$^{1,2,}$\thanks{Work done at NVIDIA. Code is available at \url{https://github.com/jeffhj/RAVEN}.} $\quad$ Wei Ping$^{2}$
$\quad$ Peng Xu$^{2}$
$\quad$ Mohammad Shoeybi$^{2}$ \\
\textbf{Kevin Chen-Chuan Chang}$^{1}$ 
$\quad$ \textbf{Bryan Catanzaro}$^{2}$ \\
$^1$University of Illinois at Urbana-Champaign $\quad$ $^2$NVIDIA \\
 \texttt{jeffhj@illinois.edu, wping@nvidia.com} \\
}

\colmfinalcopy %
\begin{document}
\maketitle
\begin{abstract}
In this paper, we investigate the in-context learning ability of retrieval-augmented encoder-decoder language models. We first conduct a comprehensive analysis of existing models and identify their limitations in in-context learning, primarily due to a mismatch between pretraining and inference, as well as a restricted context length. To address these issues, we propose \textsc{Raven}, a model that combines retrieval-augmented masked language modeling and prefix language modeling. We further introduce \textit{Fusion-in-Context Learning} to enhance the few-shot performance by enabling the model to leverage more in-context examples without requiring additional training. Through extensive experiments, we demonstrate that our simple yet effective design significantly improves performance, achieving results comparable to the most advanced language models in certain scenarios, despite having substantially fewer parameters. Our work underscores the potential of retrieval-augmented encoder-decoder language models for in-context learning and encourages further research in this direction.
\end{abstract}

\section{Introduction}

Recent advancements in natural language processing have been predominantly driven by the development of large language models (LLMs) \citep{NEURIPS2020_1457c0d6,openai2022chatgpt,openai2023gpt4,chowdhery2022palm,smith2022using}.
These models have demonstrated remarkable performance across a wide range of tasks~\citep{qin2023chatgpt,bubeck2023sparks,huang2022towards}. 
One of the key features that enables these models to excel is their ability to perform in-context learning~\citep{dong2022survey}. By conditioning on given context, LLMs can adapt to new tasks and domains without the need for task-specific fine-tuning. This enables LLMs to perform well on zero-shot or few-shot learning tasks, where only a limited number of examples are available.

While in-context learning has been extensively studied for decoder-only language models like GPT-3~\citep{NEURIPS2020_1457c0d6} and PaLM~\citep{chowdhery2022palm}, research on encoder-decoder language models, which have shown to learn stronger representations~\citep{devlin-etal-2019-bert,10.5555/3455716.3455856},  remains limited. Notably, \citet{patel2023bidirectional} tap into the potential of mT5~\citep{xue-etal-2021-mt5}, a multilingual encoder-decoder LM, by iteratively prompting the model to produce long generations with in-context examples. \citet{chung2022scaling,longpre2023flan} finetune T5~\citep{10.5555/3455716.3455856} with a large mixture of tasks using instruction tuning~\citep{mishra-etal-2022-cross,wei2022finetuned,sanh2022multitask} to improve model performance and generalization to unseen tasks in both zero-shot and few-shot settings.

On the other hand, LLMs still face challenges such as hallucination and limitations in representing the long-tail and most recent knowledge~\citep{mallen2022not,huang-etal-2022-large,luu-etal-2022-time,jang-etal-2022-temporalwiki,zheng2023does}. Retrieval-augmented language models~\citep{izacard2022few,pmlr-v162-borgeaud22a,wang2023shall,shi2023replug} have emerged as a powerful approach to address these issues by retrieving relevant knowledge from an external corpus. 
Among these, the encoder-decoder models, such as \textsc{Atlas}~\citep{izacard2022few}, stand out. They benefit from the strong representation ability of a bidirectional encoder, coupled with of the efficacy of a Fusion-in-Decoder architecture~\citep{izacard-grave-2021-leveraging}, enabling the effective integration of multiple retrieved passages.
Despite these advancements, in-context learning with these models remains underexplored.\looseness=-1

In this regard, we first conduct a comprehensive analysis of the state-of-the-art retrieval-augmented encoder-decoder language models by designing and experimenting with different prompting strategies. We find that these models exhibit a certain in-context learning ability; however, due to a mismatch between pretraining and inference and a limited context length—issues that are common to existing encoder-decoder LMs trained with masked language modeling—its few-shot performance is not stable
and providing more than, e.g., 8-shot, examples does not lead to further improvement.

Based on the analysis, we develop \textsc{Raven}\footnote{\textsc{Raven}, a bird known for its intelligence and adaptability, has the letters ``RA'' in its name, which represents ``\textbf{R}etrieval-\textbf{A}ugmented'' in our context.} by first mitigating the mismatch between pretraining and inference through a combination of retrieval-augmented masked language modeling and prefix language modeling. Moreover, to enable the model to learn from more in-context examples, we propose \textit{Fusion-in-Context Learning}, a novel approach that allows the model to utilize more in-context examples without modifying the model configuration or requiring additional training. Furthermore, we suggest using the retriever of the model to obtain relevant in-context examples to further enhance few-shot performance. Our empirical results demonstrate that \textsc{Raven} significantly outperforms previous retrieval-augmented encoder-decoder LMs in both zero-shot and few-shot settings, even achieving comparable results to decoder-only LLMs in some settings despite having 180 times fewer parameters.

The main contributions of this paper are twofold:
\begin{itemize}[leftmargin=*, nolistsep]
\setlength{\itemsep}{1mm}
\item From an analytical standpoint, we provide a thorough analysis of the in-context learning ability of retrieval-augmented encoder-decoder language models. We demonstrate the possibilities and offer insights for future development.\looseness=-1
\item From a technological perspective, we introduce \textsc{Raven}, coupled with our Fusion-in-Context Learning and In-Context Example Retrieval strategies, building upon the analytical groundwork. These techniques, though simple, are highly effective. They not only enhance the base model's capabilities but also highlight the potential of in-context learning with retrieval-augmented encoder-decoder LMs.\looseness=-1
\end{itemize}

\section{Background and Related Work}

Retrieval-augmented language models are a class of language models designed to enhance their performance by incorporating external knowledge. These models typically employ an information retrieval mechanism to access relevant information from a large corpus, which is then integrated into the model's prediction process.
Retrieval-augmented LMs can be based on both encoder-decoder~\citep{izacard2022few,NEURIPS2020_6b493230} and decoder-only~\citep{Khandelwal2020Generalization,pmlr-v162-borgeaud22a,shi-etal-2022-nearest} architectures.
For decoder-only LMs, the computational cost typically increases quadratically with the input length, as well as with the number of retrieval passages. In contrast, for encoder-decoder LMs with a Fusion-in-Decoder architecture, the computation cost grows linearly with the number of retrieved passages, as they only perform self-attention over one passage at a time \citep{izacard-grave-2021-leveraging}. This concept is also investigated by \citet{ye2023fid} for more efficient in-context learning.

While there has been some research on in-context learning with retrieval-augmented decoder-only LMs, which can be straightforwardly implemented by concatenating retrieved passages with the query as the input of the LM~\citep{mallen2022not,shi2023replug,khattab2022demonstrate}, in-context learning with retrieval-augmented encoder-decoder LMs remains unexplored to the best of our knowledge. This is despite the fact that encoder-decoder LMs can be more efficient at 
 incorporating multiple~(e.g., 40) retrieved passages.

\section{Methodology}

In this section, we first explore in-context learning with retrieval-augmented encoder-decoder language models in the literature. Building upon the analysis, we develop models with enhanced zero-shot performance and improved in-context learning abilities.

\subsection{In-Context Learning with Retrieval-Augmented Encoder-Decoder LMs}

\label{sec:analysis}

\begin{figure*}[tp!]
\centerline{\includegraphics[width=\linewidth]{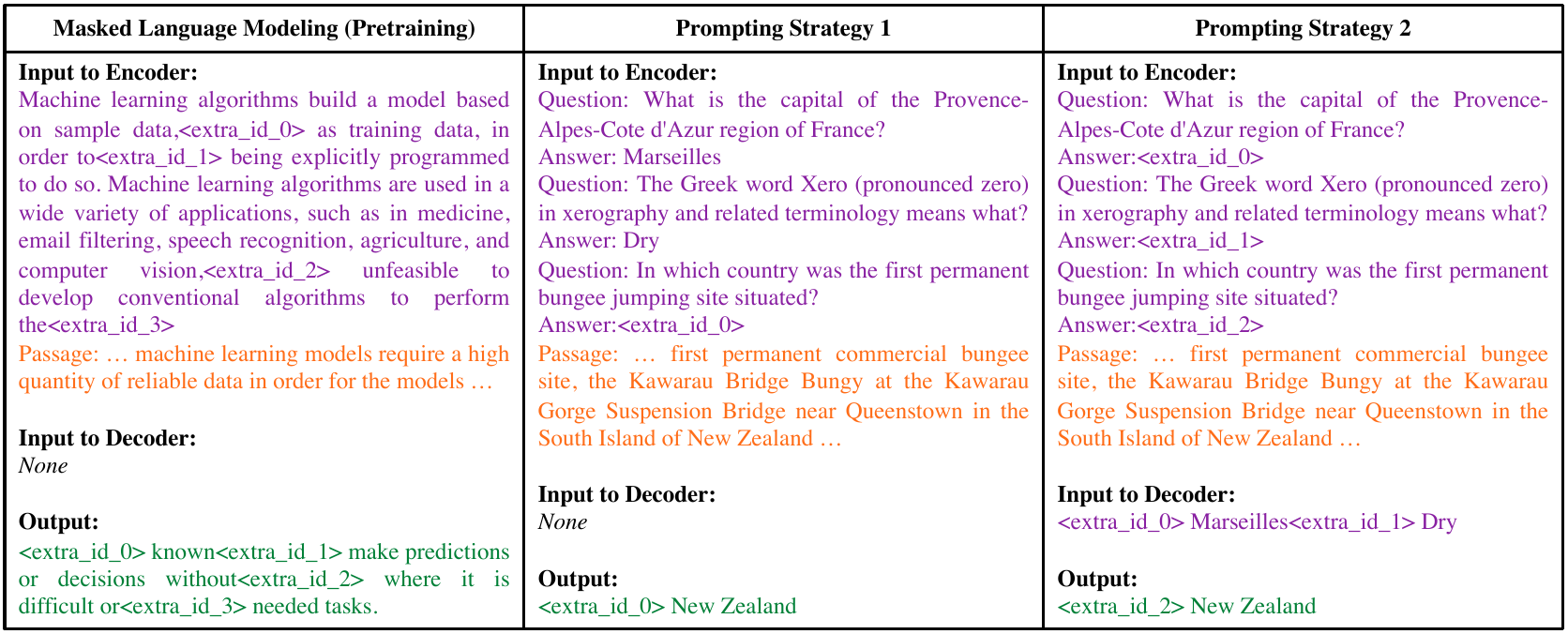}}
\vspace{-1mm}
\caption{Retrieval-augmented masked language modeling and prompting strategies for in-context learning.}
\vspace{-3mm}
\label{fig:prompting_strategy}
\end{figure*}

To investigate the in-context learning ability of retrieval-augmented encoder-decoder language models, we first aim to gain insights from the state-of-the-art designs in the literature. Among them, the design of \textsc{Atlas}~\citep{izacard2022few} stands out; it combines a general-purpose dense retriever with a sequence-to-sequence reader (i.e., T5~\citep{10.5555/3455716.3455856}) using the Fusion-in-Decoder architecture~\citep{izacard-grave-2021-leveraging}. The retriever, encoder and decoder are jointly trained during the pretraining process. In this process, the dense retriever, based on the Contriever model~\citep{izacard2022unsupervised}, is responsible for selecting relevant passages from an external knowledge source, e.g., Wikipedia, based on the given corrupted context. The retrieved passages are then processed along with the context by the encoder, which generates the corresponding output, i.e., the masked spans, at the decoder (Figure~\ref{fig:prompting_strategy}, left).
\textsc{Atlas} demonstrates exceptional few-shot performance on knowledge-intensive language tasks~\citep{petroni-etal-2021-kilt}, despite having a lower parameter count compared to other recent LLMs.\looseness=-1

However, in \citet{izacard2022few}, the few-shot performance is achieved by finetuning the model with few-shot examples, which requires additional training and may limit its applications, such as dealing with dynamic and diverse real-time user queries like GPT-3/4~\citep{NEURIPS2020_1457c0d6,openai2023gpt4}, where in-context learning plays a vital role.
Therefore, we take the initiative to explore the in-context learning ability of this type of models, using open-domain question answering~\citep{chen-etal-2017-reading} as a representative task for some preliminary experiments.

\p{Prompting Strategies.}
To facilitate in-context learning, an effective prompting strategy is paramount.
In contrast to decoder-only LMs, where the input can only be fed to the decoder, encoder-decoder LMs can take input in either the encoder or the decoder. In alignment with the pretraining objective, we identify two prompting strategies for in-context learning:

\p{Strategy 1.} The first strategy involves feeding all example question-answer pairs and the target question to the encoder, without any input to the decoder. The prompt is designed as:\footnote{Here we present a format designed for better demonstration. The actual prompt, which follows the template used in pretraining, can be found in Appendix~\ref{sec:app_openqa}.}\looseness=-1

\vspace{1mm}
\noindent\textbf{Enc}: Question: $q_1$ Answer: $a_1$ $\dots$ Question: $q_k$ Answer: $a_k$ Question: $q_0$ Answer:\mask{0} $d$ \looseness=-1

\vspace{1mm}

\noindent where $(q_1, a_1), \dots, (q_k, a_k)$ represent example QA pairs, $q_0$ denotes the target question, \mask{0} is a sentinel
token~\citep{10.5555/3455716.3455856}, and $d$ is the relevant passage retrieved with $q_0$. An example in a 2-shot setting is illusated in 
Figure~\ref{fig:prompting_strategy}~(middle).

\p{Strategy 2.} As the decoder of the encoder-decoder model can also accept input, we can feed the answers of in-context examples to the decoder and only feed the questions to the encoder, using multiple sentinel tokens:

\vspace{1mm}

\noindent \textbf{Enc}: Question: $q_1$ Answer:\mask{0} $\dots$ Question: $q_k$ Answer:\mask{$(k-1)$} Question: $q_0$ Answer:\mask{$k$} $d$

\noindent \textbf{Dec}: \mask{0} $a_1$$\dots$ \mask{$(k-1)$} $a_k$ \looseness=-1

\vspace{1mm}

Figure~\ref{fig:prompting_strategy}~(right) demonstrates an example. The model is expected to learn from in-context examples by examining both the input to the encoder and input to the decoder.

\begin{table*}[tp]
\small
\begin{center}
\begin{tabular}{ll|cccc|cccc}
\toprule
& & \multicolumn{4}{c|}{\textbf{Natural Questions}} & \multicolumn{4}{c}{\textbf{TriviaQA}} \\
& & \textbf{0-shot} & \textbf{1-shot} & \textbf{5-shot} & \textbf{8-shot} & \textbf{0-shot} & \textbf{1-shot} & \textbf{5-shot} & \textbf{8-shot} \\
\midrule 
\textsc{Atlas} & 11B S1 & \multirow{2}{*}{26.7} & 21.3 & 29.8 & \textbf{31.3} & \multirow{2}{*}{56.9} & 35.5 & 62.3 & \textbf{63.9} \\
\textsc{Atlas} & 11B S2 & & 21.4 & 16.3 & 9.8 & & 49.8 & 48.4 & 44.4 \\
\bottomrule
\end{tabular}
\end{center}
\vspace{-2mm}
\caption{Results of \textsc{Atlas} 11B with prompting strategy 1 (S1) and strategy 2 (S2).}
\label{table:result_prompting_s12}
\vspace{-1mm}
\end{table*}

We select two widely-used datasets in the domain of open-domain question answering for the preliminary study: Natural Questions (NQ)~\citep{kwiatkowski-etal-2019-natural} and TriviaQA (TQA)~\citep{joshi-etal-2017-triviaqa}\footnote{Experimental setup is detailed in the Appendix~\ref{sec:analysis_setup}.}. Table~\ref{table:result_prompting_s12} summarizes the results. We find that the model struggles to learn from in-context examples using strategy 2, as the few-shot performance is worse than the zero-shot performance.
We hypothesize that this is because the model has difficulty learning the pattern of S2 with masked language modeling during its pretraining, since it is unlikely to obtain several consecutive question-answer pairs (or something similar) in the form of strategy 2 by randomly masking several spans in a sequence.

On the other hand, we observe that with strategy 1, the model does exhibit some in-context learning ability, where the 5-shot and 8-shot performance is significantly better than the zero-shot performance on both NQ and TriviaQA. Therefore, we choose to focus on strategy 1 for further study and disregard strategy 2 for the remainder of the paper.

\begin{table}[tp]
\small
\begin{center}
\begin{tabular}{l|c|c}
\toprule
& {\textbf{Natural Questions}} & {\textbf{TriviaQA}} \\
\midrule 
\textit{first} &  0.7 & 9.2 \\
\textit{random} & 6.5 & 19.5 \\
\textit{last} & \textbf{29.8} & \textbf{62.3} \\
\bottomrule
\end{tabular}
\end{center}
\vspace{-2mm}
\caption{Results of \textsc{Atlas} 11B (5-shot) with different target question positions.}
\label{table:result_pos}
\vspace{-2mm}
\end{table}

\begin{figure}[tp]
\centerline{\includegraphics[width=0.85\linewidth]{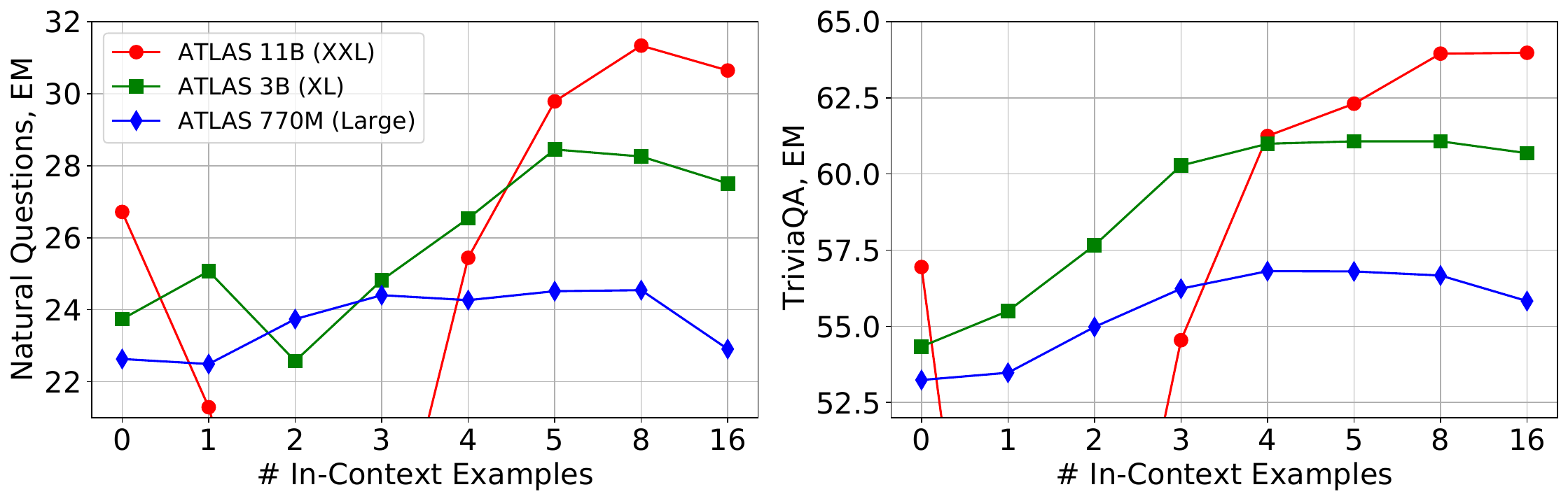}}
\vspace{-2mm}
\caption{Results of \textsc{Atlas} with different numbers of in-context examples.}
\vspace{-2mm}
\label{fig:result_num_of_ex}
\end{figure}

\p{Effect of Position.}
As the encoder of encoder-decoder language models is bidirectional, it can also examine in-context examples that follow the target question to fill in the masked token. This means that we may position the target question at the beginning or middle of a sequence, for example:\looseness=-1

\vspace{1mm}

\noindent \textcolor{violet}{Question: $q_0$ Answer:\mask{0}} Question: $q_1$ Answer: $a_1$ $\dots$ Question: $q_k$ Answer: $a_k$ $d$

\noindent Question: $q_1$ Answer: $a_1$ $\dots$ \textcolor{violet}{Question: $q_0$ Answer:\mask{0}}$\dots$ Question: $q_k$ Answer: $a_k$ $d$ \looseness=-1

\vspace{1mm}

Table~\ref{table:result_pos} summarizes the results. We denote the target question's position as ``\textit{first}'' for the beginning of the sequence, ``\textit{random}'' for a random position, and ``\textit{last}'' for the original setting (S1). Interestingly, placing the target question anywhere other than the last position results in a significant performance drop.
Upon examining the generated answers, we observe that when the target question is placed at the beginning or in the middle, the model tends to repeat the answer or generate additional text. For example, for the prompt ``Question: What number in Bingo is sometimes referred to as Heinz varieties? Answer:\mask{0} Question: \dots''. The generated text is ``57 `Heinz varieties' is a term used in Bingo to describe''. This indicates that the model does not fully understand and follow the style of in-context examples. 
Therefore, by default, we position the target question after all the in-context examples.

\p{Effect of Number of In-Context Examples.}
\label{sec:result_num_of_ex}
The number of in-context examples is a crucial hyperparameter for in-context learning. Generally, we expect better performance from a model with more in-context examples, but there is an upper limit due to 1) the maximum context length setup, e.g., 512 tokens, during the pretraining process, and 2) the point at which the model has received sufficient examples and cannot gain additional information from more examples.
The optimal number of in-context examples also varies between models. For instance, on TriviaQA, PaLM~\citep{chowdhery2022palm} exhibits better 1-shot performance than settings with more examples, while this is not the case for GPT-3~\citep{NEURIPS2020_1457c0d6}.\looseness=-1

Figure~\ref{fig:result_num_of_ex} illustrates the impact of varying the number of in-context examples across different model sizes. Interestingly, the 11B model demonstrates poor performance in low-shot settings, e.g., 1-shot, but improves significantly after 4-shot and 5-shot. Upon examining the generated responses, we find that the model tends to produce answers with more tokens in low-shot settings, while the ground truth typically consists of shorter answers with fewer than 5 tokens.
By relaxing the criteria for a correct prediction to include instances where the ground-truth answer is a substring of the model output, we find that the 1-shot performance surpasses that of the 0-shot setting (38.3 vs 32.1 on NQ).\looseness=-1

All models perform well in the 5-shot and 8-shot settings, but their performance does not continue to improve with more in-context examples (e.g., 16-shot). We believe this plateau may be attributed to two factors: 1) the sequence length constraints during pretraining, where the maximum input length to the encoder is set to 384 tokens, and the average input sequence length (excluding passages) is around 130 tokens; 2) the model's ability to learn adequately with 5 or 8 examples, making additional examples less beneficial.

\begin{figure}[tp]
\centerline{\includegraphics[width=0.85\linewidth]{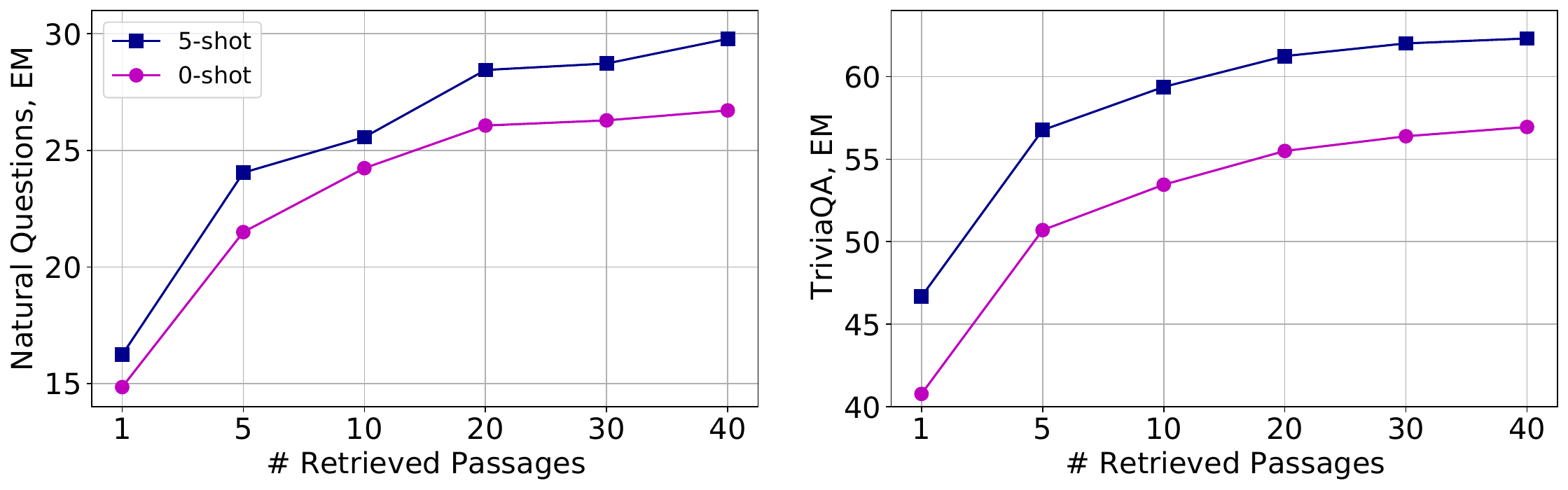}}
\vspace{-1mm}
\caption{Results of \textsc{Atlas} 11B with different numbers of retrieved passages.}
\vspace{-3mm}
\label{fig:effect_of_doc}
\end{figure}

\p{Effect of Number of Retrieved Passages.}
Figure~\ref{fig:effect_of_doc} illustrates the impact of the number of retrieved passages on model performance. We observe that for both 0-shot and 5-shot settings, the performance of the models increases significantly with the number of retrieved passages. This highlights the effectiveness of the Fusion-in-Decoder architecture~\citep{izacard-grave-2021-leveraging} for knowledge-intensive tasks like open-domain question answering, and underscores the importance of pretraining language models with retrieval augmentation.
Additionally, the 5-shot performance consistently outperforms the 0-shot setting.
This observation further emphasizes the value of providing in-context examples to improve the performance of retrieval-augmented encoder-decoder language models. \looseness-1

% In Appendices~\ref{sec:ret_position} and \ref{sec:effect_of_doc}, we also study the effect of the target question's position and the effect of the number of retrieved passages. We observe that positioning the target question after all the in-context examples yields the optimal performance. Additionally, for both 0-shot and 5-shot settings, the performance of the models increases significantly with the number of retrieved passages, with the 5-shot performance consistently outperforming the 0-shot setting. This highlights the superiority of the encoder-decoder (Fusion-in-Decoder) architecture, which offers an advantage not available to decoder-only language models.

\subsection{\textsc{Raven}: Combining Retrieval-Augmented Masked and Prefix Language Modeling}
\label{sec:raven}

\begin{wrapfigure}{R}{0.4\linewidth}
\vspace{-4.5mm}
\centerline{\includegraphics[width=\linewidth]{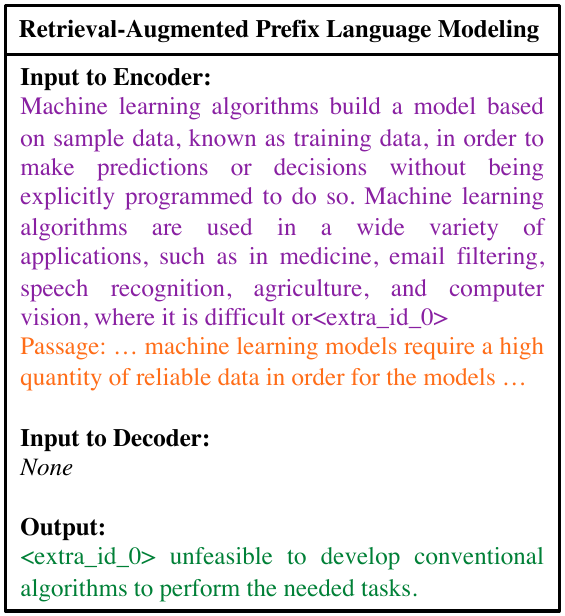}}
\vspace{-3mm}
\caption{Retrieval-augmented prefix language modeling.}
\vspace{-7mm}
\label{fig:prefix_lm}
\end{wrapfigure}

In \S\ref{sec:analysis}, we observe that retrieval-augmented encoder-decoder LMs exhibit a certain ability for in-context learning, which has been overlooked in previous studies.
However, there are also some limitations such as unstable performance in low-shot settings, and the fact that providing more in-context examples does not consistently improve performance.

To learn a better retriever and enhance the bidirectional understanding ability of the reader, as demonstrated in \citet{izacard2022few}, a practical choice is to pretrain the model with the masked language modeling objective, where the input is a corrupted text with several masked spans placed randomly within the sequence (refer to Figure~\ref{fig:prompting_strategy} (left) for an example).
However, in testing, based on our analysis in \S\ref{sec:analysis}, it is most effective to place the target question after all the in-context examples, with a masked token (i.e., \mask{0}) following the question (Figure~\ref{fig:prompting_strategy}, middle)).
Thus, there exists a mismatch between pretraining and inference.\looseness=-1

To solve this issue, we propose combining retrieval-augmented masked and prefix language modeling. Specifically, in the first stage, following \citet{izacard2022few}, the retriever and reader are trained jointly with retrieval-augmented masked language modeling. The training objective for the retriever is to minimize the KL divergence $\textsc{KL}(p_{\textsc{reader}} \ \| \ p_{\textsc{retriever}})$ between the passage posterior distribution according to the reader and the passage distribution from the retriever over the top-K retrieved passages, i.e.,
$
p_{\textsc{reader}}({d}) = \frac{\exp(\log p_{LM} ({a} \ | \ {d}, {q}))}{\sum_{i=1}^K \exp ( \log p_{LM} ({a} \ | \ {d}_i, {q}))},\ 
p_{\textsc{retriever}}({d}) = \frac{\exp(s({d}, {q}) / T)}{\sum_{i=1}^K \exp(s({d}_i, {q}) / T)},
$
where $s(\cdot)$ calculates the dot product between the query $q$ and passage $d$ vectors, and $T$ is a hyperparameter. The training objective for the reader is to maximize the likelihood of the masked spans with $n$ retrieved passages: $\sum_{i} \log p({a}_i\ | \ {q}, \{d_k\}_{1,\dots,n}, {a}_{1:i-1})$.

In the second stage, for each sequence, we mask 10\% of the tokens on average at the end of the sequence with the \mask{0} token. Then, we use the retriever obtained from the first stage to retrieve relevant passages using the prefix and train the reader to recover the suffix of this sequence with the prefix and the passages as input. An example of input and output for retrieval-augmented prefix language modeling is shown in Figure~\ref{fig:prefix_lm}. We can observe that the pretraining objective aligns more closely with prompting strategy 1 in Figure~\ref{fig:prompting_strategy}. We refer to the model trained with this combined objective as \textsc{Raven}. 

\textsc{Raven} benefits from both the retrieval-augmented masked language modeling, which contributes to a better reader and retriever, and retrieval-augmented prefix language modeling, which mitigates the gap between pretraining and inference. This design is \textit{non-trivial}. In Appendix~\ref{sec:ablation}, 
we verify the effectiveness of it by exploring different training strategies.

\subsection{Fusion-in-Context Learning}

In \S\ref{sec:result_num_of_ex}, we observe that the performance does not further improve with more in-context examples after 8-shot. One major reason for this is the limited sequence length during the pretraining process, which makes it difficult for the model to handle long sequences during inference. Pretraining models with longer contexts would be a potential solution, but it would significantly increase computation cost. Additionally, the maximum input length is also constrained by the maximum sequence length of the retriever, i.e., Contriever, which is based on BERT~\citep{devlin-etal-2019-bert} and has a maximum length of 512 tokens.

\begin{figure}
\centerline{\includegraphics[width=0.9\linewidth]{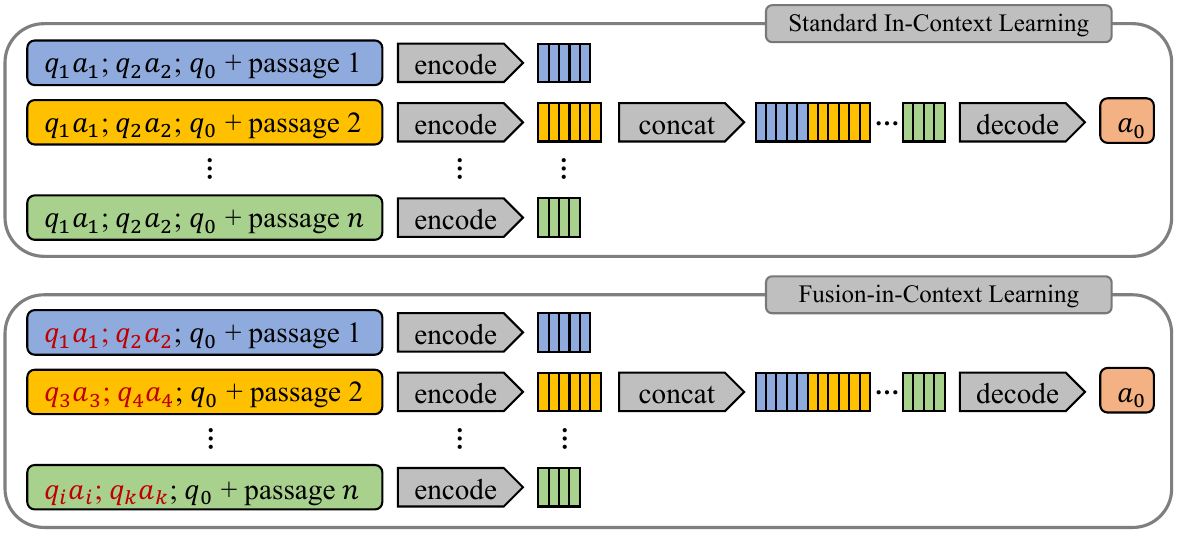}}
\vspace{-1mm}
\caption{Standard In-Context Learning vs Fusion-in-Context Learning.}
\vspace{-3mm}
\label{fig:FiCL}
\end{figure}

As an alternative, we propose an approach to enable models to learn from more in-context examples without requiring additional training. As described in \S\ref{sec:analysis}, the reader is based on the Fusion-in-Decoder architecture~\citep{izacard-grave-2021-leveraging}, where multiple passages are retrieved, and each passage, concatenated with the in-context examples and target question, is fed to the encoder separately (Figure~\ref{fig:FiCL}, top).
To allow the model to process more in-context examples, we can feed \textit{different} in-context examples to the encoder with each passage (Figure~\ref{fig:FiCL}, bottom). In this way, the model can incorporate more in-context examples during its inference process. We refer to this strategy as \textit{Fusion-in-Context Learning (FiCL)}.\looseness=-1

In implementation, for a $k$-shot setting, such as a 64-shot setting, to effectively utilize the 64 examples, we randomly shuffle these examples and select $m$ (e.g., 5) examples in order as the input for the encoder each time.
If all the examples have been used, we shuffle the 64 examples again. We denote the configuration of FiCL as [$k$-$m$], which stands for [$k$-shot, $m$-fusion].\looseness=-1

\subsection{In-Context Example Retrieval}
\label{sec:ex_re}

\citet{liu-etal-2022-makes,rubin-etal-2022-learning,su2023selective} demonstrate that a well-chosen selection of in-context examples can enhance in-context learning. Building on this insight, we propose utilizing the retriever of \textsc{Raven} to retrieve in-context examples.
Specifically, we use \textsc{Raven}'s retriever to build an index during the preparation step, and then, during testing, when the model receives an input, it could efficiently retrieve in-context examples with its retriever.

By integrating \textsc{Raven}'s retriever in this manner, we aim to: 1) automate in-context learning, which is particularly practical for model owners who have a database of examples. Without this, users would need to manually provide in-context examples; and 2) optimize the selection of in-context examples, thereby improving in-context learning performance.

\section{Experiments}
\subsection{Experimental Setup}

\p{Datasets.} Following the setup in \S\ref{sec:analysis}, we first evaluate on two widely-used open-domain question answering datasets: Natural Questions~\citep{kwiatkowski-etal-2019-natural} and TriviaQA~\citep{joshi-etal-2017-triviaqa}. 
Additionally, we conduct a case study on long-form question answering using the ELI5 dataset~\citep{fan-etal-2019-eli5}.
Furthermore, we assess the models' language understanding ability using the Massively Multitask Language Understanding (MMLU) benchmark \citep{hendrycks2021measuring}. 
Detailed information regarding the MMLU evaluation is in Appendix~\ref{sec:details_mmlu}.
Other evaluation settings are the same as \S\ref{sec:analysis_setup}. 

\p{Baselines.} Since both \textsc{Raven} and \textsc{Atlas}~\citep{izacard2022few} are trained starting from T5, we choose \textsc{Atlas} as a primary baseline for comparison. We also compare our model with decoder-only LLMs such as GPT-3~\citep{NEURIPS2020_1457c0d6}, PaLM~\citep{chowdhery2022palm}, and LLaMA~\citep{touvron2023llama} (in a closed-book setting). Additionally, for open-domain QA, we evaluate our approach against \textsc{RePlug}~\citep{shi2023replug} and \textsc{Retro}~\citep{pmlr-v162-borgeaud22a}, as well as its improved version \textsc{Retro++}~\citep{wang2023shall}. These models are decoder-only language models augmented with retrieval.
\textsc{RePlug} is based on Codex~\citep{chen2021evaluating} and Contriever~\citep{izacard2022unsupervised}, where the passages are retrieved by Contriever (using ensemble and additional adaptation) and fed directly to Codex. \textsc{Retro} is a GPT model~\citep{radford2019language} augmented with a transformer encoder to encode the retrieved passages. \textsc{Retro++} is a variant of \textsc{Retro} that feeds the most relevant retrieved passage into the GPT decoder while providing other passages to its encoder.
For MMLU, we also include FLAN-T5~\citep{chung2022scaling}, an enhanced version of T5 that has been trained on a large mixture of tasks with instruction finetuning.\footnote{Implementation details are described in Appendix~\ref{sec:training_details}.}

\begin{figure*}[tp!]
\centerline{\includegraphics[width=0.85\linewidth]{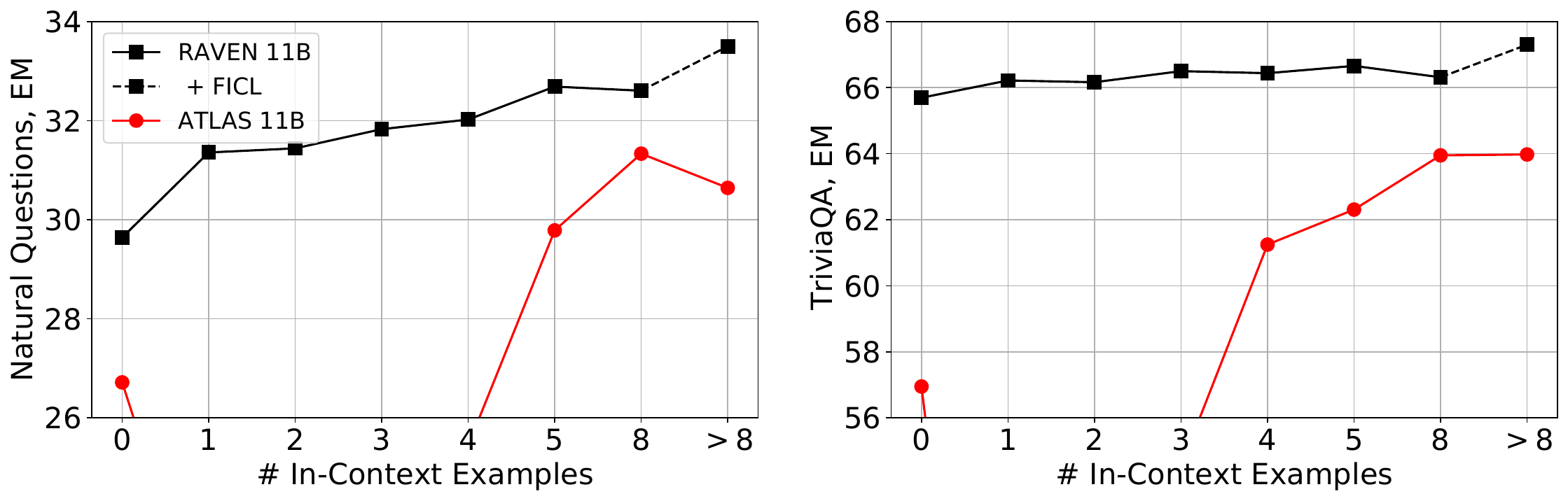}}
\vspace{-2mm}
\caption{\textsc{Raven} vs \textsc{Atlas}. We report the best observed performance achieved with more than eight shots for ``$>8$''.}
\label{fig:result_raven_atlas}
\vspace{-3mm}
\end{figure*}

\begin{table*}[ht]
\small
\begin{center}
\begin{tabular}{ll|ccl|ccl}
\toprule
& & \multicolumn{3}{c|}{\textbf{Natural Questions}} & \multicolumn{3}{c}{\textbf{TriviaQA}} \\
& & \textbf{0-shot} & \textbf{1-shot} & \textbf{few-shot} & \textbf{0-shot} & \textbf{1-shot} & \textbf{few-shot} \\
\midrule 
GPT-3 & 13B & \ 7.8 & 13.7 & 21.0 \tiny{(64)} & 41.8 & 51.3 & 57.5 \tiny{(64)} \\
GPT-3 & 175B & 14.6 & 23.0 & 29.9 \tiny{(64)} & 64.3 & 68.0 & 71.2 \tiny{(64)} \\
PaLM & 8B & \ 8.4 & 10.6 & 14.6 \tiny{(5)} & 39.5 & 48.5 & 47.2 \tiny{(5)} \\
PaLM & 62B & 18.1 & 23.1 & 27.6 \tiny{(5)} & 67.3 & 72.7 & 70.1 \tiny{(5)}\\
PaLM & 540B & 21.2 & 29.3 & 39.6 \tiny{(64)} & \textbf{76.9} & \textbf{81.4} & \textbf{81.4} \tiny{(1)}* \\
Codex & 175B & - & - & 40.6 \tiny{(16)} & - & - & 73.6 \tiny{(16)} \\
LLaMA &  7B & 16.8 & 18.7 & 26.1 \tiny{(64)} & 50.0 & 53.4 & 57.6 \tiny{(64)} \\
LLaMA &  65B & 23.8 & 31.0 & 39.9 \tiny{(64)} & 68.2 &  71.6 & 73.0 \tiny{(64)} \\
\midrule
\midrule
\multicolumn{8}{l}{\textbf{Retrieval-Augmented Language Models}} \\
\midrule
Codex + Contriever & 175B & - & - & 44.2 \tiny{(16)} & - & - & 76.0 \tiny{(16)}\\
Codex + \textsc{RePlug} & 175B & - & - & 44.7 \tiny{(16)} & - & - & 76.8 \tiny{(16)} \\
Codex + \textsc{RePlug} LSR & 175B & - & - & \textbf{45.5} \tiny{(16)} & - & - & 77.3 \tiny{(16)} \\
\textsc{Retro} & 9.5B & 8.9 & - & \ \ \ - & 36.0 & - & \ \ \ - \\
\textsc{Retro++} & 9.5B & 25.8 & - & \ \ \ - & 48.3 & - & \ \ \ - \\
\hline
\textsc{Atlas} & 3B & 23.7 & 25.1 & 28.4 \tiny{(5)} & 54.3 & 55.5 & 61.1 \tiny{(5)} \\
\textsc{Atlas} + FiCL & 3B & & & {29.6 \tiny{[64-5]}} & & & {62.0 \tiny{[64-5]}} \\
\textsc{Atlas} & 11B & 26.7 & 21.3 & 31.3 \tiny{(8)} & 56.9 & 35.5 & 63.9 \tiny{(8)} \\
\textsc{Atlas} + FiCL & 11B & & & 32.0 \tiny{[64-8]} & & & {64.9 \tiny{[64-8]}} \\
\hline
\textbf{\textsc{Raven}} & 3B & 29.3 & \textbf{31.7} & 31.4 \tiny{(5)} & 62.4 & 63.2 & 62.6 \tiny{(5)} \\
\textbf{\textsc{Raven}} + FiCL & 3B & & & {32.8 \tiny{[40-1]}} & & & {63.6 \tiny{[40-1]}} \\
\textbf{\textsc{Raven}} & 11B & \textbf{29.6} & 31.4 & 32.7 \tiny{(5)} & {65.7} & {66.2} & 66.7 \tiny{(5)} \\
\textbf{\textsc{Raven}} + FiCL & 11B & & & {33.5 \tiny{[64-5]}} & & & {{67.3} \tiny{[64-5]}} \\
\bottomrule
\multicolumn{8}{l}{\scriptsize{* For TriviaQA, PaLM's 1-shot performance surpasses other settings. We follow the original paper to report the 1-shot result.}} \\
\multicolumn{8}{l}{\scriptsize{\ \ \ For other models, we select the best $k$-shot ($k \in \{2,3,4,5,8,16\}$) performance or report the number in the original paper.}} \\
\end{tabular}
\end{center}
\vspace{-2mm}
\caption{Results on NQ and TriviaQA. Since the performance varies significantly depending on the capability of the base model, the results from models other than \textsc{Atlas} should only be used for reference to gauge the position. And we assume \textsc{Raven} can achieve significant performance improvement when based on a stronger base model.}
\label{table:main_result}
\end{table*}

\subsection{Open-Domain Question Answering}
\label{sec:exp_openqa}

We choose open-domain QA as our primary evaluation task, as it effectively represents knowledge-intensive challenges and is widely employed in real-world applications.

\p{\textsc{Raven} vs \textsc{Atlas}.}
Figure~\ref{fig:result_raven_atlas} and Table~\ref{table:main_result} present the exact match (EM) scores for \textsc{Atlas} and \textsc{Raven} on the NQ and TriviaQA datasets. Both the 3B and 11B \textsc{Raven} models significantly outperform \textsc{Atlas}. 
For instance, on TriviaQA, \textsc{Raven} 11B achieves an improvement of 8.8\%, 30.7\%, and 2.8\% in the 0-shot, 1-shot, and few-shot settings respectively, compared to \textsc{Atlas} 11B.
Furthermore, the performance of \textsc{Raven} increases steadily with the number of in-context examples, while the performance of \textsc{Atlas} experiences a substantial decline in low-shot settings, demonstrating the effectiveness of \textsc{Raven} across various shot settings.

\p{Fusion-in-Context Learning.} We also report the results of models with Fusion-in-Context Learning (FiCL) in Table~\ref{table:main_result}. For both \textsc{Atlas} and \textsc{Raven}, FiCL contributes to approximately a 1\% improvement, which is not attainable by standard in-context learning, where performance does not further improve (or even decreases) with more than $8$ in-context examples. This demonstrates the superiority of FiCL for enabling models to learn from more examples.\looseness=-1

\begin{figure*}[tp]
\centerline{\includegraphics[width=0.85\linewidth]{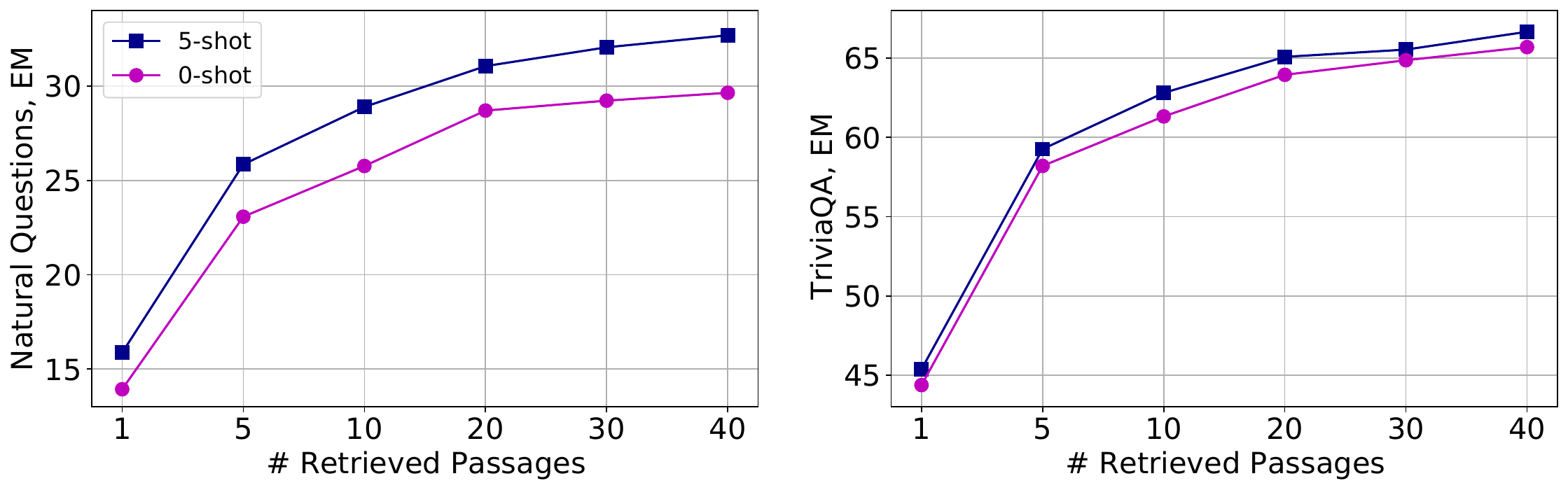}}
\vspace{-2mm}
\caption{Results of \textsc{Raven} 11B with different numbers of retrieved passages.}
\vspace{-3mm}
\label{fig:effect_of_doc_raven}
\end{figure*}

\p{Comparison to Other Models.} In Table~\ref{table:main_result}, we further compare \textsc{Raven} to other baselines.
On NQ, \textsc{Raven}'s zero-shot and one-shot performance surpasses all the baselines, including PaLM, even though \textsc{Raven} 3B has 180 times fewer parameters than PaLM 540B. The zero-shot performance of \textsc{Raven} on TriviaQA is also on par with PaLM 62B.
Furthermore, \textsc{Raven}'s zero-shot performance significantly exceeds that of both \textsc{Retro} and \textsc{Retro++}, which are retrieval-augmented language models of a similar scale.

In the few-shot setting, with FiCL, \textsc{Raven} achieves performance comparable to GPT-3 175B and PaLM 62B. However, there remains a gap between \textsc{Raven} and the larger PaLM 540B and Codex 175B models. Nevertheless, given the considerably smaller scale of \textsc{Raven} in comparison to PaLM and Codex, its performance can be considered impressive. The performance of \textsc{Raven} may be further improved if it is built upon a larger model, in which case its few-shot performance is likely to surpass that of PaLM and Codex.

% \begin{wraptable}{R}{0.47\textwidth}
\begin{table}[tp]
% \vspace{38mm}
\small
\begin{center}
\begin{tabular}{l|cc|cc}
\toprule
& \multicolumn{2}{c|}{\textbf{NQ}} & \multicolumn{2}{c}{\textbf{TQA}} \\
& \textbf{1-shot} & \textbf{5-shot} & \textbf{1-shot} & \textbf{5-shot} \\
\midrule 
 3B & +9.1 & +11.6 & +0.0 & +1.6 \\
 11B & +9.8 & +11.1 & -0.5 & +1.0 \\
\bottomrule
\end{tabular}
\end{center}
\vspace{-2mm}
\caption{Performance improvement of \textsc{Raven} with In-Context Example Retrieval.}
\label{table:result_ex_re}
\vspace{-2mm}
% \end{wraptable}
\end{table}

\p{Effect of Number of Retrieved Passages.}
Figure~\ref{fig:effect_of_doc_raven} illustrates the effect of the number of retrieved passages. As the number of retrieved passages increases, we observe a significant performance improvement of \textsc{Raven} 11B in both the 0-shot and 5-shot settings.

\p{In-Context Example Retrieval.}
\S\ref{sec:ex_re} suggests using \textsc{Raven}'s retriever for in-context example retrieval. Results in Table~\ref{table:result_ex_re} show that this approach improves \textsc{Raven}'s few-shot results, especially on NQ where a $\sim$10\% improvement is observed. This indicates the positive impact of incorporating more relevant in-context examples.

\p{Additional Results.} We conduct an \textbf{ablation study of different training strategies} in Appendix~\ref{sec:ablation} and provide a \textbf{case study on long-form question answering} in Appendix~\ref{sec:eli5}.

\subsection{MMLU}

\begin{wraptable}{R}{0.5\textwidth}
\vspace{-5mm}
\small
\begin{center}
\begin{tabular}{ll|ccl}
\toprule
& & \textbf{0-shot} & \textbf{1-shot} & \textbf{5-shot} \\
\midrule
GPT-3 & 13B & - & - & 26.0 \\
GPT-3 & 175B & - & - & 43.9 \\
PaLM & 8B & - & - & 25.3 \\
PaLM & 62B & - & - & 53.7 \\
PaLM & 540B & - & - & 69.3 \\
\midrule
T5 & 3B & - & - & 25.7 \\
T5 & 11B & - & - & 25.9 \\
FLAN-T5 & 3B & - & - & 52.4 \\
FLAN-T5 & 11B & - & - & 55.1 \\
\midrule
\textsc{Atlas} & 3B & 43.7 & 36.9 & 38.5 \\
\rowcolor{gray!20}
\ + FiCL & 3B & & & 42.6 \tiny{[40-1]}   \\
\textsc{Atlas} & 11B & 47.4 & 45.3 & 44.2 \\
\rowcolor{gray!20}
 \ + FiCL & 11B & & & 48.0 \tiny{[40-1]}  \\
\textsc{Raven} & 3B & 45.7 & 40.0 & 40.4  \\
\rowcolor{gray!20}
 \ + FiCL & 3B & & & 44.5 \tiny{[64-5]}  \\
\textsc{Raven} & 11B & 48.9 & 49.2 & 48.7  \\
\rowcolor{gray!20}
 \ + FiCL & 11B & & & 50.5 \tiny{[40-1]} \\
\bottomrule
\end{tabular}
\end{center}
\vspace{-2mm}
\caption{Results on MMLU.}
\label{table:result_mmlu}
\vspace{-3mm}
\end{wraptable}

Table~\ref{table:result_mmlu} summarizes the results (accuracy) on Massive Multitask Language Understanding (MMLU). We find that the zero-shot performance of \textsc{Raven} is impressive, surpassing the few-shot performance of GPT-3 175B and being slightly worse than PaLM 62B, despite having a significantly smaller number of parameters. Furthermore, with the same number of parameters, the performance of \textsc{Raven} is far superior to T5.
Additionally, even without instruction finetuning, \textsc{Raven} achieves performance comparable to FLAN-T5, a model finetuned on a large collection of tasks. We expect further improvement of \textsc{Raven} by applying instruction tuning as well and leave it for future study.

Interestingly, with standard in-context learning, the few-shot performance of \textsc{Raven} is worse than zero-shot, possibly due to the longer questions and answer options in MMLU causing context length issues in the 5-shot setting. 
Also, in the one-shot setting, since MMLU is a multiple-choice QA task, providing only one example might introduce bias in the model's prediction, favoring a specific option. However, with Fusion-in-Context Learning, the performance improves significantly, leading to better few-shot performance for the 11B model compared to its zero-shot performance, further demonstrating the effectiveness of FiCL.\looseness=-1

\section{Conclusion}

In this study, we have delved into the in-context learning ability of retrieval-augmented encoder-decoder language models. We commenced with a comprehensive analysis of the models in the literature and subsequently developed our model based on the analysis.
Our extensive experimental results demonstrated that our model significantly outperforms previous models and achieves results on par with some of the most advanced language models, even with substantially fewer parameters.
These findings highlight the potential of retrieval-augmented encoder-decoder language models in the realm of in-context learning.

\bibliography{colm2024_conference}

\begin{thebibliography}{48}
\providecommand{\natexlab}[1]{#1}
\providecommand{\url}[1]{\texttt{#1}}
\expandafter\ifx\csname urlstyle\endcsname\relax
  \providecommand{\doi}[1]{doi: #1}\else
  \providecommand{\doi}{doi: \begingroup \urlstyle{rm}\Url}\fi

\bibitem[Borgeaud et~al.(2022)Borgeaud, Mensch, Hoffmann, Cai, Rutherford, Millican, Van Den~Driessche, Lespiau, Damoc, Clark, De~Las~Casas, Guy, Menick, Ring, Hennigan, Huang, Maggiore, Jones, Cassirer, Brock, Paganini, Irving, Vinyals, Osindero, Simonyan, Rae, Elsen, and Sifre]{pmlr-v162-borgeaud22a}
Sebastian Borgeaud, Arthur Mensch, Jordan Hoffmann, Trevor Cai, Eliza Rutherford, Katie Millican, George~Bm Van Den~Driessche, Jean-Baptiste Lespiau, Bogdan Damoc, Aidan Clark, Diego De~Las~Casas, Aurelia Guy, Jacob Menick, Roman Ring, Tom Hennigan, Saffron Huang, Loren Maggiore, Chris Jones, Albin Cassirer, Andy Brock, Michela Paganini, Geoffrey Irving, Oriol Vinyals, Simon Osindero, Karen Simonyan, Jack Rae, Erich Elsen, and Laurent Sifre.
\newblock Improving language models by retrieving from trillions of tokens.
\newblock In Kamalika Chaudhuri, Stefanie Jegelka, Le~Song, Csaba Szepesvari, Gang Niu, and Sivan Sabato (eds.), \emph{Proceedings of the 39th International Conference on Machine Learning}, volume 162 of \emph{Proceedings of Machine Learning Research}, pp.\  2206--2240. PMLR, 17--23 Jul 2022.
\newblock URL \url{https://proceedings.mlr.press/v162/borgeaud22a.html}.

\bibitem[Brown et~al.(2020)Brown, Mann, Ryder, Subbiah, Kaplan, Dhariwal, Neelakantan, Shyam, Sastry, Askell, Agarwal, Herbert-Voss, Krueger, Henighan, Child, Ramesh, Ziegler, Wu, Winter, Hesse, Chen, Sigler, Litwin, Gray, Chess, Clark, Berner, McCandlish, Radford, Sutskever, and Amodei]{NEURIPS2020_1457c0d6}
Tom Brown, Benjamin Mann, Nick Ryder, Melanie Subbiah, Jared~D Kaplan, Prafulla Dhariwal, Arvind Neelakantan, Pranav Shyam, Girish Sastry, Amanda Askell, Sandhini Agarwal, Ariel Herbert-Voss, Gretchen Krueger, Tom Henighan, Rewon Child, Aditya Ramesh, Daniel Ziegler, Jeffrey Wu, Clemens Winter, Chris Hesse, Mark Chen, Eric Sigler, Mateusz Litwin, Scott Gray, Benjamin Chess, Jack Clark, Christopher Berner, Sam McCandlish, Alec Radford, Ilya Sutskever, and Dario Amodei.
\newblock Language models are few-shot learners.
\newblock In H.~Larochelle, M.~Ranzato, R.~Hadsell, M.F. Balcan, and H.~Lin (eds.), \emph{Advances in Neural Information Processing Systems}, volume~33, pp.\  1877--1901. Curran Associates, Inc., 2020.
\newblock URL \url{https://proceedings.neurips.cc/paper_files/paper/2020/file/1457c0d6bfcb4967418bfb8ac142f64a-Paper.pdf}.

\bibitem[Bubeck et~al.(2023)Bubeck, Chandrasekaran, Eldan, Gehrke, Horvitz, Kamar, Lee, Lee, Li, Lundberg, et~al.]{bubeck2023sparks}
S{\'e}bastien Bubeck, Varun Chandrasekaran, Ronen Eldan, Johannes Gehrke, Eric Horvitz, Ece Kamar, Peter Lee, Yin~Tat Lee, Yuanzhi Li, Scott Lundberg, et~al.
\newblock Sparks of artificial general intelligence: Early experiments with gpt-4.
\newblock \emph{arXiv preprint arXiv:2303.12712}, 2023.

\bibitem[Chen et~al.(2017)Chen, Fisch, Weston, and Bordes]{chen-etal-2017-reading}
Danqi Chen, Adam Fisch, Jason Weston, and Antoine Bordes.
\newblock Reading {W}ikipedia to answer open-domain questions.
\newblock In \emph{Proceedings of the 55th Annual Meeting of the Association for Computational Linguistics (Volume 1: Long Papers)}, pp.\  1870--1879, Vancouver, Canada, July 2017. Association for Computational Linguistics.
\newblock \doi{10.18653/v1/P17-1171}.
\newblock URL \url{https://aclanthology.org/P17-1171}.

\bibitem[Chen et~al.(2021)Chen, Tworek, Jun, Yuan, Pinto, Kaplan, Edwards, Burda, Joseph, Brockman, et~al.]{chen2021evaluating}
Mark Chen, Jerry Tworek, Heewoo Jun, Qiming Yuan, Henrique Ponde de~Oliveira Pinto, Jared Kaplan, Harri Edwards, Yuri Burda, Nicholas Joseph, Greg Brockman, et~al.
\newblock Evaluating large language models trained on code.
\newblock \emph{arXiv preprint arXiv:2107.03374}, 2021.
\newblock URL \url{https://arxiv.org/abs/2107.03374}.

\bibitem[Chowdhery et~al.(2023)Chowdhery, Narang, Devlin, Bosma, Mishra, Roberts, Barham, Chung, Sutton, Gehrmann, Schuh, Shi, Tsvyashchenko, Maynez, Rao, Barnes, Tay, Shazeer, Prabhakaran, Reif, Du, Hutchinson, Pope, Bradbury, Austin, Isard, Gur-Ari, Yin, Duke, Levskaya, Ghemawat, Dev, Michalewski, Garcia, Misra, Robinson, Fedus, Zhou, Ippolito, Luan, Lim, Zoph, Spiridonov, Sepassi, Dohan, Agrawal, Omernick, Dai, Pillai, Pellat, Lewkowycz, Moreira, Child, Polozov, Lee, Zhou, Wang, Saeta, Diaz, Firat, Catasta, Wei, Meier-Hellstern, Eck, Dean, Petrov, and Fiedel]{chowdhery2022palm}
Aakanksha Chowdhery, Sharan Narang, Jacob Devlin, Maarten Bosma, Gaurav Mishra, Adam Roberts, Paul Barham, Hyung~Won Chung, Charles Sutton, Sebastian Gehrmann, Parker Schuh, Kensen Shi, Sasha Tsvyashchenko, Joshua Maynez, Abhishek Rao, Parker Barnes, Yi~Tay, Noam Shazeer, Vinodkumar Prabhakaran, Emily Reif, Nan Du, Ben Hutchinson, Reiner Pope, James Bradbury, Jacob Austin, Michael Isard, Guy Gur-Ari, Pengcheng Yin, Toju Duke, Anselm Levskaya, Sanjay Ghemawat, Sunipa Dev, Henryk Michalewski, Xavier Garcia, Vedant Misra, Kevin Robinson, Liam Fedus, Denny Zhou, Daphne Ippolito, David Luan, Hyeontaek Lim, Barret Zoph, Alexander Spiridonov, Ryan Sepassi, David Dohan, Shivani Agrawal, Mark Omernick, Andrew~M. Dai, Thanumalayan~Sankaranarayana Pillai, Marie Pellat, Aitor Lewkowycz, Erica Moreira, Rewon Child, Oleksandr Polozov, Katherine Lee, Zongwei Zhou, Xuezhi Wang, Brennan Saeta, Mark Diaz, Orhan Firat, Michele Catasta, Jason Wei, Kathy Meier-Hellstern, Douglas Eck, Jeff Dean, Slav Petrov, and Noah Fiedel.
\newblock Palm: Scaling language modeling with pathways.
\newblock \emph{Journal of Machine Learning Research}, 24\penalty0 (240):\penalty0 1--113, 2023.
\newblock URL \url{http://jmlr.org/papers/v24/22-1144.html}.

\bibitem[Chung et~al.(2022)Chung, Hou, Longpre, Zoph, Tay, Fedus, Li, Wang, Dehghani, Brahma, et~al.]{chung2022scaling}
Hyung~Won Chung, Le~Hou, Shayne Longpre, Barret Zoph, Yi~Tay, William Fedus, Eric Li, Xuezhi Wang, Mostafa Dehghani, Siddhartha Brahma, et~al.
\newblock Scaling instruction-finetuned language models.
\newblock \emph{arXiv preprint arXiv:2210.11416}, 2022.
\newblock URL \url{https://arxiv.org/abs/2210.11416}.

\bibitem[Devlin et~al.(2019)Devlin, Chang, Lee, and Toutanova]{devlin-etal-2019-bert}
Jacob Devlin, Ming-Wei Chang, Kenton Lee, and Kristina Toutanova.
\newblock {BERT}: Pre-training of deep bidirectional transformers for language understanding.
\newblock In \emph{Proceedings of the 2019 Conference of the North {A}merican Chapter of the Association for Computational Linguistics: Human Language Technologies, Volume 1 (Long and Short Papers)}, pp.\  4171--4186, Minneapolis, Minnesota, June 2019. Association for Computational Linguistics.
\newblock \doi{10.18653/v1/N19-1423}.
\newblock URL \url{https://aclanthology.org/N19-1423}.

\bibitem[Dong et~al.(2022)Dong, Li, Dai, Zheng, Wu, Chang, Sun, Xu, and Sui]{dong2022survey}
Qingxiu Dong, Lei Li, Damai Dai, Ce~Zheng, Zhiyong Wu, Baobao Chang, Xu~Sun, Jingjing Xu, and Zhifang Sui.
\newblock A survey for in-context learning.
\newblock \emph{arXiv preprint arXiv:2301.00234}, 2022.
\newblock URL \url{https://arxiv.org/abs/2301.00234}.

\bibitem[Fan et~al.(2019)Fan, Jernite, Perez, Grangier, Weston, and Auli]{fan-etal-2019-eli5}
Angela Fan, Yacine Jernite, Ethan Perez, David Grangier, Jason Weston, and Michael Auli.
\newblock {ELI}5: Long form question answering.
\newblock In \emph{Proceedings of the 57th Annual Meeting of the Association for Computational Linguistics}, pp.\  3558--3567, Florence, Italy, July 2019. Association for Computational Linguistics.
\newblock \doi{10.18653/v1/P19-1346}.
\newblock URL \url{https://aclanthology.org/P19-1346}.

\bibitem[Hendrycks et~al.(2021)Hendrycks, Burns, Basart, Zou, Mazeika, Song, and Steinhardt]{hendrycks2021measuring}
Dan Hendrycks, Collin Burns, Steven Basart, Andy Zou, Mantas Mazeika, Dawn Song, and Jacob Steinhardt.
\newblock Measuring massive multitask language understanding.
\newblock In \emph{International Conference on Learning Representations}, 2021.
\newblock URL \url{https://openreview.net/forum?id=d7KBjmI3GmQ}.

\bibitem[Huang \& Chang(2023)Huang and Chang]{huang2022towards}
Jie Huang and Kevin Chen-Chuan Chang.
\newblock Towards reasoning in large language models: A survey.
\newblock In \emph{Findings of the Association for Computational Linguistics: ACL 2023}, pp.\  1049--1065, Toronto, Canada, July 2023. Association for Computational Linguistics.
\newblock \doi{10.18653/v1/2023.findings-acl.67}.
\newblock URL \url{https://aclanthology.org/2023.findings-acl.67}.

\bibitem[Huang et~al.(2022)Huang, Shao, and Chang]{huang-etal-2022-large}
Jie Huang, Hanyin Shao, and Kevin Chen-Chuan Chang.
\newblock Are large pre-trained language models leaking your personal information?
\newblock In \emph{Findings of the Association for Computational Linguistics: EMNLP 2022}, pp.\  2038--2047, Abu Dhabi, United Arab Emirates, December 2022. Association for Computational Linguistics.
\newblock URL \url{https://aclanthology.org/2022.findings-emnlp.148}.

\bibitem[Izacard \& Grave(2021)Izacard and Grave]{izacard-grave-2021-leveraging}
Gautier Izacard and Edouard Grave.
\newblock Leveraging passage retrieval with generative models for open domain question answering.
\newblock In \emph{Proceedings of the 16th Conference of the European Chapter of the Association for Computational Linguistics: Main Volume}, pp.\  874--880, Online, April 2021. Association for Computational Linguistics.
\newblock \doi{10.18653/v1/2021.eacl-main.74}.
\newblock URL \url{https://aclanthology.org/2021.eacl-main.74}.

\bibitem[Izacard et~al.(2022)Izacard, Caron, Hosseini, Riedel, Bojanowski, Joulin, and Grave]{izacard2022unsupervised}
Gautier Izacard, Mathilde Caron, Lucas Hosseini, Sebastian Riedel, Piotr Bojanowski, Armand Joulin, and Edouard Grave.
\newblock Unsupervised dense information retrieval with contrastive learning.
\newblock \emph{Transactions on Machine Learning Research}, 2022.
\newblock ISSN 2835-8856.
\newblock URL \url{https://openreview.net/forum?id=jKN1pXi7b0}.

\bibitem[Izacard et~al.(2023)Izacard, Lewis, Lomeli, Hosseini, Petroni, Schick, Dwivedi-Yu, Joulin, Riedel, and Grave]{izacard2022few}
Gautier Izacard, Patrick Lewis, Maria Lomeli, Lucas Hosseini, Fabio Petroni, Timo Schick, Jane Dwivedi-Yu, Armand Joulin, Sebastian Riedel, and Edouard Grave.
\newblock Atlas: Few-shot learning with retrieval augmented language models.
\newblock \emph{Journal of Machine Learning Research}, 24\penalty0 (251):\penalty0 1--43, 2023.
\newblock URL \url{http://jmlr.org/papers/v24/23-0037.html}.

\bibitem[Jang et~al.(2022)Jang, Ye, Lee, Yang, Shin, Han, Kim, and Seo]{jang-etal-2022-temporalwiki}
Joel Jang, Seonghyeon Ye, Changho Lee, Sohee Yang, Joongbo Shin, Janghoon Han, Gyeonghun Kim, and Minjoon Seo.
\newblock {T}emporal{W}iki: A lifelong benchmark for training and evaluating ever-evolving language models.
\newblock In \emph{Proceedings of the 2022 Conference on Empirical Methods in Natural Language Processing}, pp.\  6237--6250, Abu Dhabi, United Arab Emirates, December 2022. Association for Computational Linguistics.
\newblock URL \url{https://aclanthology.org/2022.emnlp-main.418}.

\bibitem[Joshi et~al.(2017)Joshi, Choi, Weld, and Zettlemoyer]{joshi-etal-2017-triviaqa}
Mandar Joshi, Eunsol Choi, Daniel Weld, and Luke Zettlemoyer.
\newblock {T}rivia{QA}: A large scale distantly supervised challenge dataset for reading comprehension.
\newblock In \emph{Proceedings of the 55th Annual Meeting of the Association for Computational Linguistics (Volume 1: Long Papers)}, pp.\  1601--1611, Vancouver, Canada, July 2017. Association for Computational Linguistics.
\newblock \doi{10.18653/v1/P17-1147}.
\newblock URL \url{https://aclanthology.org/P17-1147}.

\bibitem[Khandelwal et~al.(2020)Khandelwal, Levy, Jurafsky, Zettlemoyer, and Lewis]{Khandelwal2020Generalization}
Urvashi Khandelwal, Omer Levy, Dan Jurafsky, Luke Zettlemoyer, and Mike Lewis.
\newblock Generalization through memorization: Nearest neighbor language models.
\newblock In \emph{International Conference on Learning Representations}, 2020.
\newblock URL \url{https://openreview.net/forum?id=HklBjCEKvH}.

\bibitem[Khattab et~al.(2022)Khattab, Santhanam, Li, Hall, Liang, Potts, and Zaharia]{khattab2022demonstrate}
Omar Khattab, Keshav Santhanam, Xiang~Lisa Li, David Hall, Percy Liang, Christopher Potts, and Matei Zaharia.
\newblock Demonstrate-search-predict: Composing retrieval and language models for knowledge-intensive nlp.
\newblock \emph{arXiv preprint arXiv:2212.14024}, 2022.

\bibitem[Kwiatkowski et~al.(2019)Kwiatkowski, Palomaki, Redfield, Collins, Parikh, Alberti, Epstein, Polosukhin, Devlin, Lee, Toutanova, Jones, Kelcey, Chang, Dai, Uszkoreit, Le, and Petrov]{kwiatkowski-etal-2019-natural}
Tom Kwiatkowski, Jennimaria Palomaki, Olivia Redfield, Michael Collins, Ankur Parikh, Chris Alberti, Danielle Epstein, Illia Polosukhin, Jacob Devlin, Kenton Lee, Kristina Toutanova, Llion Jones, Matthew Kelcey, Ming-Wei Chang, Andrew~M. Dai, Jakob Uszkoreit, Quoc Le, and Slav Petrov.
\newblock Natural questions: A benchmark for question answering research.
\newblock \emph{Transactions of the Association for Computational Linguistics}, 7:\penalty0 452--466, 2019.
\newblock \doi{10.1162/tacl_a_00276}.
\newblock URL \url{https://aclanthology.org/Q19-1026}.

\bibitem[Lewis et~al.(2020)Lewis, Perez, Piktus, Petroni, Karpukhin, Goyal, K\"{u}ttler, Lewis, Yih, Rockt\"{a}schel, Riedel, and Kiela]{NEURIPS2020_6b493230}
Patrick Lewis, Ethan Perez, Aleksandra Piktus, Fabio Petroni, Vladimir Karpukhin, Naman Goyal, Heinrich K\"{u}ttler, Mike Lewis, Wen-tau Yih, Tim Rockt\"{a}schel, Sebastian Riedel, and Douwe Kiela.
\newblock Retrieval-augmented generation for knowledge-intensive nlp tasks.
\newblock In H.~Larochelle, M.~Ranzato, R.~Hadsell, M.F. Balcan, and H.~Lin (eds.), \emph{Advances in Neural Information Processing Systems}, volume~33, pp.\  9459--9474. Curran Associates, Inc., 2020.
\newblock URL \url{https://proceedings.neurips.cc/paper_files/paper/2020/file/6b493230205f780e1bc26945df7481e5-Paper.pdf}.

\bibitem[Liu et~al.(2022)Liu, Shen, Zhang, Dolan, Carin, and Chen]{liu-etal-2022-makes}
Jiachang Liu, Dinghan Shen, Yizhe Zhang, Bill Dolan, Lawrence Carin, and Weizhu Chen.
\newblock What makes good in-context examples for {GPT}-3?
\newblock In \emph{Proceedings of Deep Learning Inside Out (DeeLIO 2022): The 3rd Workshop on Knowledge Extraction and Integration for Deep Learning Architectures}, pp.\  100--114, Dublin, Ireland and Online, May 2022. Association for Computational Linguistics.
\newblock \doi{10.18653/v1/2022.deelio-1.10}.
\newblock URL \url{https://aclanthology.org/2022.deelio-1.10}.

\bibitem[Longpre et~al.(2023)Longpre, Hou, Vu, Webson, Chung, Tay, Zhou, Le, Zoph, Wei, et~al.]{longpre2023flan}
Shayne Longpre, Le~Hou, Tu~Vu, Albert Webson, Hyung~Won Chung, Yi~Tay, Denny Zhou, Quoc~V Le, Barret Zoph, Jason Wei, et~al.
\newblock The flan collection: Designing data and methods for effective instruction tuning.
\newblock \emph{arXiv preprint arXiv:2301.13688}, 2023.
\newblock URL \url{https://arxiv.org/abs/2301.13688}.

\bibitem[Loshchilov \& Hutter(2019)Loshchilov and Hutter]{loshchilov2018decoupled}
Ilya Loshchilov and Frank Hutter.
\newblock Decoupled weight decay regularization.
\newblock In \emph{International Conference on Learning Representations}, 2019.
\newblock URL \url{https://openreview.net/forum?id=Bkg6RiCqY7}.

\bibitem[Luu et~al.(2022)Luu, Khashabi, Gururangan, Mandyam, and Smith]{luu-etal-2022-time}
Kelvin Luu, Daniel Khashabi, Suchin Gururangan, Karishma Mandyam, and Noah~A. Smith.
\newblock Time waits for no one! analysis and challenges of temporal misalignment.
\newblock In \emph{Proceedings of the 2022 Conference of the North American Chapter of the Association for Computational Linguistics: Human Language Technologies}, pp.\  5944--5958, Seattle, United States, July 2022. Association for Computational Linguistics.
\newblock \doi{10.18653/v1/2022.naacl-main.435}.
\newblock URL \url{https://aclanthology.org/2022.naacl-main.435}.

\bibitem[Mallen et~al.(2022)Mallen, Asai, Zhong, Das, Hajishirzi, and Khashabi]{mallen2022not}
Alex Mallen, Akari Asai, Victor Zhong, Rajarshi Das, Hannaneh Hajishirzi, and Daniel Khashabi.
\newblock When not to trust language models: Investigating effectiveness and limitations of parametric and non-parametric memories.
\newblock \emph{arXiv preprint arXiv:2212.10511}, 2022.
\newblock URL \url{https://arxiv.org/abs/2212.10511}.

\bibitem[Mishra et~al.(2022)Mishra, Khashabi, Baral, and Hajishirzi]{mishra-etal-2022-cross}
Swaroop Mishra, Daniel Khashabi, Chitta Baral, and Hannaneh Hajishirzi.
\newblock Cross-task generalization via natural language crowdsourcing instructions.
\newblock In \emph{Proceedings of the 60th Annual Meeting of the Association for Computational Linguistics (Volume 1: Long Papers)}, pp.\  3470--3487, Dublin, Ireland, May 2022. Association for Computational Linguistics.
\newblock \doi{10.18653/v1/2022.acl-long.244}.
\newblock URL \url{https://aclanthology.org/2022.acl-long.244}.

\bibitem[OpenAI(2022)]{openai2022chatgpt}
OpenAI.
\newblock Chatgpt: Optimizing language models for dialogue.
\newblock \emph{OpenAI}, 2022.

\bibitem[OpenAI(2023)]{openai2023gpt4}
OpenAI.
\newblock Gpt-4 technical report, 2023.

\bibitem[Patel et~al.(2023)Patel, Li, Rasooli, Constant, Raffel, and Callison-Burch]{patel2023bidirectional}
Ajay Patel, Bryan Li, Mohammad~Sadegh Rasooli, Noah Constant, Colin Raffel, and Chris Callison-Burch.
\newblock Bidirectional language models are also few-shot learners.
\newblock In \emph{The Eleventh International Conference on Learning Representations}, 2023.
\newblock URL \url{https://openreview.net/forum?id=wCFB37bzud4}.

\bibitem[Petroni et~al.(2021)Petroni, Piktus, Fan, Lewis, Yazdani, De~Cao, Thorne, Jernite, Karpukhin, Maillard, Plachouras, Rockt{\"a}schel, and Riedel]{petroni-etal-2021-kilt}
Fabio Petroni, Aleksandra Piktus, Angela Fan, Patrick Lewis, Majid Yazdani, Nicola De~Cao, James Thorne, Yacine Jernite, Vladimir Karpukhin, Jean Maillard, Vassilis Plachouras, Tim Rockt{\"a}schel, and Sebastian Riedel.
\newblock {KILT}: a benchmark for knowledge intensive language tasks.
\newblock In \emph{Proceedings of the 2021 Conference of the North American Chapter of the Association for Computational Linguistics: Human Language Technologies}, pp.\  2523--2544, Online, June 2021. Association for Computational Linguistics.
\newblock \doi{10.18653/v1/2021.naacl-main.200}.
\newblock URL \url{https://aclanthology.org/2021.naacl-main.200}.

\bibitem[Qin et~al.(2023)Qin, Zhang, Zhang, Chen, Yasunaga, and Yang]{qin2023chatgpt}
Chengwei Qin, Aston Zhang, Zhuosheng Zhang, Jiaao Chen, Michihiro Yasunaga, and Diyi Yang.
\newblock Is chatgpt a general-purpose natural language processing task solver?
\newblock \emph{arXiv preprint arXiv:2302.06476}, 2023.
\newblock URL \url{https://arxiv.org/abs/2302.06476}.

\bibitem[Radford et~al.(2019)Radford, Wu, Child, Luan, Amodei, Sutskever, et~al.]{radford2019language}
Alec Radford, Jeffrey Wu, Rewon Child, David Luan, Dario Amodei, Ilya Sutskever, et~al.
\newblock Language models are unsupervised multitask learners.
\newblock \emph{OpenAI blog}, 1\penalty0 (8):\penalty0 9, 2019.

\bibitem[Raffel et~al.(2020)Raffel, Shazeer, Roberts, Lee, Narang, Matena, Zhou, Li, and Liu]{10.5555/3455716.3455856}
Colin Raffel, Noam Shazeer, Adam Roberts, Katherine Lee, Sharan Narang, Michael Matena, Yanqi Zhou, Wei Li, and Peter~J. Liu.
\newblock Exploring the limits of transfer learning with a unified text-to-text transformer.
\newblock \emph{J. Mach. Learn. Res.}, 21\penalty0 (1), jan 2020.
\newblock ISSN 1532-4435.
\newblock URL \url{https://dl.acm.org/doi/abs/10.5555/3455716.3455856}.

\bibitem[Rubin et~al.(2022)Rubin, Herzig, and Berant]{rubin-etal-2022-learning}
Ohad Rubin, Jonathan Herzig, and Jonathan Berant.
\newblock Learning to retrieve prompts for in-context learning.
\newblock In \emph{Proceedings of the 2022 Conference of the North American Chapter of the Association for Computational Linguistics: Human Language Technologies}, pp.\  2655--2671, Seattle, United States, July 2022. Association for Computational Linguistics.
\newblock \doi{10.18653/v1/2022.naacl-main.191}.
\newblock URL \url{https://aclanthology.org/2022.naacl-main.191}.

\bibitem[Sanh et~al.(2022)Sanh, Webson, Raffel, Bach, Sutawika, Alyafeai, Chaffin, Stiegler, Raja, Dey, Bari, Xu, Thakker, Sharma, Szczechla, Kim, Chhablani, Nayak, Datta, Chang, Jiang, Wang, Manica, Shen, Yong, Pandey, Bawden, Wang, Neeraj, Rozen, Sharma, Santilli, Fevry, Fries, Teehan, Scao, Biderman, Gao, Wolf, and Rush]{sanh2022multitask}
Victor Sanh, Albert Webson, Colin Raffel, Stephen Bach, Lintang Sutawika, Zaid Alyafeai, Antoine Chaffin, Arnaud Stiegler, Arun Raja, Manan Dey, M~Saiful Bari, Canwen Xu, Urmish Thakker, Shanya~Sharma Sharma, Eliza Szczechla, Taewoon Kim, Gunjan Chhablani, Nihal Nayak, Debajyoti Datta, Jonathan Chang, Mike Tian-Jian Jiang, Han Wang, Matteo Manica, Sheng Shen, Zheng~Xin Yong, Harshit Pandey, Rachel Bawden, Thomas Wang, Trishala Neeraj, Jos Rozen, Abheesht Sharma, Andrea Santilli, Thibault Fevry, Jason~Alan Fries, Ryan Teehan, Teven~Le Scao, Stella Biderman, Leo Gao, Thomas Wolf, and Alexander~M Rush.
\newblock Multitask prompted training enables zero-shot task generalization.
\newblock In \emph{International Conference on Learning Representations}, 2022.
\newblock URL \url{https://openreview.net/forum?id=9Vrb9D0WI4}.

\bibitem[Shi et~al.(2022)Shi, Michael, Gururangan, and Zettlemoyer]{shi-etal-2022-nearest}
Weijia Shi, Julian Michael, Suchin Gururangan, and Luke Zettlemoyer.
\newblock Nearest neighbor zero-shot inference.
\newblock In \emph{Proceedings of the 2022 Conference on Empirical Methods in Natural Language Processing}, pp.\  3254--3265, Abu Dhabi, United Arab Emirates, December 2022. Association for Computational Linguistics.
\newblock URL \url{https://aclanthology.org/2022.emnlp-main.214}.

\bibitem[Shi et~al.(2023)Shi, Min, Yasunaga, Seo, James, Lewis, Zettlemoyer, and Yih]{shi2023replug}
Weijia Shi, Sewon Min, Michihiro Yasunaga, Minjoon Seo, Rich James, Mike Lewis, Luke Zettlemoyer, and Wen-tau Yih.
\newblock Replug: Retrieval-augmented black-box language models.
\newblock \emph{arXiv preprint arXiv:2301.12652}, 2023.
\newblock URL \url{https://arxiv.org/abs/2301.12652}.

\bibitem[Smith et~al.(2022)Smith, Patwary, Norick, LeGresley, Rajbhandari, Casper, Liu, Prabhumoye, Zerveas, Korthikanti, et~al.]{smith2022using}
Shaden Smith, Mostofa Patwary, Brandon Norick, Patrick LeGresley, Samyam Rajbhandari, Jared Casper, Zhun Liu, Shrimai Prabhumoye, George Zerveas, Vijay Korthikanti, et~al.
\newblock Using deepspeed and megatron to train megatron-turing nlg 530b, a large-scale generative language model.
\newblock \emph{arXiv preprint arXiv:2201.11990}, 2022.
\newblock URL \url{https://arxiv.org/abs/2201.11990}.

\bibitem[Su et~al.(2023)Su, Kasai, Wu, Shi, Wang, Xin, Zhang, Ostendorf, Zettlemoyer, Smith, and Yu]{su2023selective}
Hongjin Su, Jungo Kasai, Chen~Henry Wu, Weijia Shi, Tianlu Wang, Jiayi Xin, Rui Zhang, Mari Ostendorf, Luke Zettlemoyer, Noah~A. Smith, and Tao Yu.
\newblock Selective annotation makes language models better few-shot learners.
\newblock In \emph{The Eleventh International Conference on Learning Representations}, 2023.
\newblock URL \url{https://openreview.net/forum?id=qY1hlv7gwg}.

\bibitem[Tay et~al.(2023)Tay, Dehghani, Tran, Garcia, Wei, Wang, Chung, Bahri, Schuster, Zheng, Zhou, Houlsby, and Metzler]{tay2023ul}
Yi~Tay, Mostafa Dehghani, Vinh~Q. Tran, Xavier Garcia, Jason Wei, Xuezhi Wang, Hyung~Won Chung, Dara Bahri, Tal Schuster, Steven Zheng, Denny Zhou, Neil Houlsby, and Donald Metzler.
\newblock {UL}2: Unifying language learning paradigms.
\newblock In \emph{The Eleventh International Conference on Learning Representations}, 2023.
\newblock URL \url{https://openreview.net/forum?id=6ruVLB727MC}.

\bibitem[Touvron et~al.(2023)Touvron, Lavril, Izacard, Martinet, Lachaux, Lacroix, Rozi{\`e}re, Goyal, Hambro, Azhar, et~al.]{touvron2023llama}
Hugo Touvron, Thibaut Lavril, Gautier Izacard, Xavier Martinet, Marie-Anne Lachaux, Timoth{\'e}e Lacroix, Baptiste Rozi{\`e}re, Naman Goyal, Eric Hambro, Faisal Azhar, et~al.
\newblock Llama: Open and efficient foundation language models.
\newblock \emph{arXiv preprint arXiv:2302.13971}, 2023.
\newblock URL \url{https://arxiv.org/abs/2302.13971}.

\bibitem[Wang et~al.(2023)Wang, Ping, Xu, McAfee, Liu, Shoeybi, Dong, Kuchaiev, Li, Xiao, et~al.]{wang2023shall}
Boxin Wang, Wei Ping, Peng Xu, Lawrence McAfee, Zihan Liu, Mohammad Shoeybi, Yi~Dong, Oleksii Kuchaiev, Bo~Li, Chaowei Xiao, et~al.
\newblock Shall we pretrain autoregressive language models with retrieval? a comprehensive study.
\newblock In \emph{Proceedings of the 2023 Conference on Empirical Methods in Natural Language Processing}, pp.\  7763--7786, 2023.
\newblock URL \url{https://aclanthology.org/2023.emnlp-main.482}.

\bibitem[Wei et~al.(2022)Wei, Bosma, Zhao, Guu, Yu, Lester, Du, Dai, and Le]{wei2022finetuned}
Jason Wei, Maarten Bosma, Vincent Zhao, Kelvin Guu, Adams~Wei Yu, Brian Lester, Nan Du, Andrew~M. Dai, and Quoc~V Le.
\newblock Finetuned language models are zero-shot learners.
\newblock In \emph{International Conference on Learning Representations}, 2022.
\newblock URL \url{https://openreview.net/forum?id=gEZrGCozdqR}.

\bibitem[Xue et~al.(2021)Xue, Constant, Roberts, Kale, Al-Rfou, Siddhant, Barua, and Raffel]{xue-etal-2021-mt5}
Linting Xue, Noah Constant, Adam Roberts, Mihir Kale, Rami Al-Rfou, Aditya Siddhant, Aditya Barua, and Colin Raffel.
\newblock m{T}5: A massively multilingual pre-trained text-to-text transformer.
\newblock In \emph{Proceedings of the 2021 Conference of the North American Chapter of the Association for Computational Linguistics: Human Language Technologies}, pp.\  483--498, Online, June 2021. Association for Computational Linguistics.
\newblock \doi{10.18653/v1/2021.naacl-main.41}.
\newblock URL \url{https://aclanthology.org/2021.naacl-main.41}.

\bibitem[Ye et~al.(2023)Ye, Beltagy, Peters, Ren, and Hajishirzi]{ye2023fid}
Qinyuan Ye, Iz~Beltagy, Matthew~E Peters, Xiang Ren, and Hannaneh Hajishirzi.
\newblock Fid-icl: A fusion-in-decoder approach for efficient in-context learning.
\newblock In \emph{Proceedings of the 61st Annual Meeting of the Association for Computational Linguistics (Volume 1: Long Papers)}, pp.\  8158--8185, 2023.
\newblock URL \url{https://aclanthology.org/2023.acl-long.454}.

\bibitem[Zheng et~al.(2023)Zheng, Huang, and Chang]{zheng2023does}
Shen Zheng, Jie Huang, and Kevin Chen-Chuan Chang.
\newblock Why does {ChatGPT} fall short in providing truthful answers?
\newblock In \emph{I Can't Believe It's Not Better Workshop: Failure Modes in the Age of Foundation Models}, 2023.
\newblock URL \url{https://openreview.net/forum?id=w7o14LCw9P}.

\end{thebibliography}
\bibliographystyle{colm2024_conference}

\clearpage

\appendix

\section{Limitations and Broader Impact}

While the performance of \textsc{Raven} is impressive considering its scale and training budget, there are also some limitations. One limitation arises from the constrained context length inherent to the base models (i.e., T5 or \textsc{Atlas}) we employed. This restriction poses challenges to the scalability of in-context learning, especially as the number of in-context examples increases. While our Fusion-in-Context Learning (FiCL) strategy does offer a mitigative approach to this constraint, an alternative and possibly more optimal solution might involve extending the context length. This would be particularly beneficial for tasks requiring extensive inputs.\looseness=-1

Furthermore, when compared to some of the prevailing decoder-only language models, particularly those exceeding 100B parameters, the models deployed in our research might appear relatively diminutive in scale (in terms of both the number of parameters and the amount of training data). 
Our endeavor partially seeks to catalyze further investigations into more powerful encoder-decoder models.\looseness=-1

Nonetheless, the insights and methods proposed are transferable and have the potential to enhance other models, including those that are domain-specialized or more powerful, such as mT5 \citep{xue-etal-2021-mt5} and UL2 \citep{tay2023ul}.
Future work focusing on scaling up the model, applying these methods, and further studying its in-context learning ability is encouraged.
Drawing on the benefits of scaling up and combining this with our proposed approaches, we believe that there is potential to develop even more powerful retrieval-augmented language models in the future.
Another promising future direction is exploring how to combine the Fusion-in-Decoder architecture with existing decoder-only language models. By doing so, we can harness the advantages of both architectures—employing a bidirectional architecture to effectively encode retrieved passages for the most powerful decoder-only LLMs.

\section{Additional Experimental Details}
\label{sec:additional_details}

\subsection{Experimental Setup for \S\ref{sec:analysis}}
\label{sec:analysis_setup}

We select two widely-used datasets in the domain of open-domain question answering: Natural Questions (NQ)~\citep{kwiatkowski-etal-2019-natural} and TriviaQA (TQA)~\citep{joshi-etal-2017-triviaqa}.
To assess the performance, we follow the previous work~\citep{izacard2022few} to employ the standard exact match (EM) metric.
For the few-shot settings, we follow \citet{NEURIPS2020_1457c0d6} to evaluate each example in the test set by generating in-context examples through randomly sampling $k$ instances from the respective task's training set.
Following \citet{izacard2022few}, we use an index composed of December 2018 Wikipedia dump for NQ and an index composed of December 2021 Wikipedia corpora for TriviaQA.
We retrieve 40 documents by default.
We test the checkpoints released in the official repository of \citet{izacard2022few}\footnote{\url{https://github.com/facebookresearch/atlas}}, covering sizes of 11B (XXL), 3B (XL), and 770M (Large).

\subsection{Training Details}
\label{sec:training_details}

We train two versions of \textsc{Raven}: 3B and 11B. To isolate the effect of training variance with masked language modeling, we initialize both the retriever and the reader of the models with the weights of \textsc{Atlas} (3B and 11B) and continue to pretrain the model with prefix language modeling. To isolate the effect of retrieval, we do not update the retriever during the training process for prefix language modeling. We pretrain the reader using the December 2021 Wikipedia corpora preprocessed by \citet{izacard2022few}, where the index is also constructed using the same corpora. In accordance with \citet{izacard2022few}, we retrieve 20 passages for each masked sequence (excluding passages identical to the original sequence). Both the 3B and 11B models are trained for 5,000 steps, using AdamW optimizer~\citep{loshchilov2018decoupled} with a batch size of 64. We employ a learning rate of $4 \times 10^{-5}$ for the 3B model and $1 \times 10^{-5}$ for the 11B model, with linear decay and 100 warmup steps. All the models are trained on NVIDIA A100 GPUs (80 GB).
For the 3B model, we utilize 8 GPUs, whereas for the 11B model, we employ 32 GPUs. The prompt used for prefix language modeling is detailed in Appendix~\ref{sec:app_prefix_lm}. During testing, we default to retrieving 40 documents for all tasks. The prompts used can be found in Appendix~\ref{sec:app_openqa} and Appendix~\ref{sec:details_mmlu}.

\subsection{Retrieval-Augmented Prefix Language Modeling}
\label{sec:app_prefix_lm}

In alignment with the pretraining of \textsc{Atlas}, we design the prompt for prefix language modeling as 

\vspace{1mm}

\noindent \texttt{\{prefix\}\mask{0} title: \{title\} context: \{text\}}

\vspace{1mm}

\noindent where \{prefix\} represents the prefix of an input sequence. The \{title\} and \{text\} elements are retrieved by the model's retriever using the prefix as a query. Here, \{text\} signifies the retrieved passage, while \{title\} denotes the corresponding article and section title of the passage.
The model is trained to generate 

\vspace{1mm}

\noindent \texttt{\mask{0}\{suffix\}}

\vspace{1mm}

\noindent where \{suffix\} is the suffix (masked by \mask{0}) of the input sequence.

\subsection{Open-Domain Question Answering}
\label{sec:app_openqa}

In accordance with pretraining, we use the following prompt for open-domain question answering:

\vspace{1mm}

\noindent \texttt{Question: \{question\} Answer:\mask{0} title: \{title\} context: \{text\}}

\vspace{1mm}

For example, 

\vspace{1mm}

\noindent \texttt{Question: In which country was the first permanent bungee jumping site situated? Answer:\mask{0} title: Bungee jumping: Modern sport context: first permanent commercial bungee site, the Kawarau Bridge Bungy at the Kawarau Gorge Suspension Bridge near Queenstown in the South Island of New Zealand. Hackett remains one of the largest commercial operators, with concerns in several countries. Several million successful jumps have taken place since 1980. This safety record is attributable to bungee operators rigorously conforming to standards and guidelines governing jumps, such as double checking calculations and fittings for every jump. As with any sport, injuries can still occur (see below), and there have been fatalities. A relatively common mistake in fatality cases is to use a cord that}

\subsection{MMLU}
\label{sec:details_mmlu}

MMLU comprises 57 multiple-choice question answering datasets that span various domains, including elementary mathematics, US history, computer science, and more.
For the evaluation on MMLU, we report the accuracy and use an index composed of December 2021 Wikipedia corpora. 
We follow \citet{izacard2022few} to apply the ``de-biased'' inference.
Specifically, during inference, we execute four forward passes, each corresponding to a cyclic permutation of the answer letter-option assignment within the question. For instance, the answer option designated to letter `A' is shifted to `B', `B' to `C', `C' to `D', and `D' to `A'. The final prediction is obtained by summing up the probabilities from these four forward passes.

We design the prompt in the following format:

\vspace{1mm}

\noindent \texttt{Question: \{question\} Options: \{candidate answers\} Answer:\mask{0} title: \{title\} context: \{text\}}

\vspace{1mm}

For example, 

\vspace{1mm}

\noindent \texttt{Question: Over time, non-volcanic mountains can form due to the interaction of plate boundaries. Which interaction is most likely associated with the formation of non-volcanic mountains? Options: (A) continental plates colliding with continental plates (B) continental plates separating from continental plates (C) oceanic plates colliding with oceanic plates (D) oceanic plates separating from oceanic plates Answer:\mask{0} title: ... context: ...}

\vspace{1mm}

Given that many questions in the MMLU benchmark are quite lengthy, concatenating in-context examples (questions and candidate answers) with the target question in a few-shot setting is likely to exceed the maximum input length. To mitigate this, we only sample examples with question lengths of fewer than 50 tokens to use as in-context examples.

\section{Additional Results}

\subsection{Ablation Study}
\label{sec:ablation}

\begin{table*}[tp]
\small
\begin{center}
\begin{tabular}{ll|ccc|ccc}
\toprule
& & \multicolumn{3}{c|}{\textbf{Natural Questions}} & \multicolumn{3}{c}{\textbf{TriviaQA}} \\
& & \textbf{0-shot} & \textbf{1-shot} & \textbf{5-shot} & \textbf{0-shot} & \textbf{1-shot} & \textbf{5-shot} \\
\midrule 
\textsc{Atlas} & 3B (Mask) & 23.7 & 25.1 & 28.4  & 54.3 & 55.5 & 61.1 \\
\textsc{Atlas} & 3B (Mask, 5k more steps) & 22.9 & 22.5 & 28.1 & 50.8 & 50.1 & 61.1 \\
\midrule
\textsc{Raven}$^-$ & 3B (Prefix) & 24.8 & 29.1 & 30.1 & 55.4 & 61.4 & 62.3 \\
\textsc{Raven}$^-$ & 3B (Mix) & 25.1 & 28.4 & 30.9 & 56.1 & 61.4 & 62.2 \\
\midrule
\textsc{Raven} & 3B & 29.3 & 31.7 & 31.4  & 62.4 & 63.2 & 62.6 \\
\bottomrule
\end{tabular}
\end{center}
\vspace{-2mm}
\caption{Results of \textsc{Atlas} and \textsc{Raven} trained with different strategies.}
\label{table:result_ablation}
\vspace{-3mm}
\end{table*}

We conduct an ablation study by training \textsc{Atlas} and \textsc{Raven} with different pretraining strategies. First, to isolate the effect of more training steps of \textsc{Raven}, we also train \textsc{Atlas} for 5,000 more steps using the masked language modeling objective. Results in Table~\ref{table:result_ablation} (row 2) show that the performance does not improve, indicating that the performance improvement of \textsc{Raven} compared to \textsc{Atlas} is not simply due to training for more steps.

Second, to verify the effectiveness of \textsc{Raven}'s training strategy (i.e., first masked language modeling, and then prefix language modeling), we train two variants of \textsc{Raven}, starting from the T5-lm-adapt checkpoint\footnote{\url{https://huggingface.co/google/t5-xl-lm-adapt}}, which is the checkpoint that \textsc{Atlas} starts from. For the first variant, we use the same prefix language modeling objective of \textsc{Raven}. For the second variant, we train the model with a mixture of masked and prefix language modeling. Specifically, we construct corrupted texts by both masking 15\% spans in the sequence (same as \textsc{Atlas}) and replacing the suffix with a special mask token \mask{99} (used in testing). We train the model for 10,000 steps and update the retriever and refresh the index during training with the optimal strategy described in \citet{izacard2022few}. Table~\ref{table:result_ablation} (\textsc{Raven}$^-$ in row 3 and 4) summarizes the results. We find that the performance of these two variants is superior to \textsc{Atlas}, but inferior to \textsc{Raven} when trained using the strategy described in \S\ref{sec:raven}.
An explanation for this is that, by training with masked language modeling first, the model can achieve better language understanding ability and is equipped with a more effective retriever (as emperically verified in \citet{izacard2022few}). Subsequently, by training with prefix language modeling, the mismatch between pretraining and inference is mitigated, resulting in improved zero-shot and few-shot performance.

\subsection{Long-Form Question Answering}
\label{sec:eli5}

Table~\ref{table:eli5} presents some example outputs of \textsc{Atlas} and \textsc{Raven} 11B on long-form question answering. The questions are sampled from the ELI5 dataset~\citep{fan-etal-2019-eli5}.
An examination of these results reveals that \textsc{Atlas} typically generates concise answers, while the output from \textsc{Raven} generally encompasses more information. This is a predictable outcome given that \textsc{Atlas} is pretrained solely with masked language modeling, where each masked span usually contains only a handful of tokens. Besides, while \textsc{Raven}'s answers are not always entirely accurate, they generally exhibit higher quality compared to \textsc{Atlas}. Furthermore, the use of Fusion-in-Context Learning in \textsc{Raven} appears to contribute to a more coherent and informative generation.

\begin{table*}[tp]
\centering
\small
\begin{tabular}{ll|p{0.77\linewidth}}
\toprule
\multicolumn{2}{l|}{\textbf{Question}} &  \textbf{How does a painting like ``When Will You Marry'' sell for \$300 million dollars when it doesn't look like anything spectacular} \\
\midrule
\multicolumn{2}{l|}{Human}  & When it comes to a lot of art, context is more important than the actual material. Rarity, historical context etc. make a thing more valuable than its actual aesthetic properties. For example, some shitty buggy SNES or N64 games will sell for huge amounts if they are well preserved and in the original packaging etc, even when any high school student nowadays could write a better game in a week. \\
\hline
\textsc{Atlas} & 1-shot & It's a masterpiece. \\
\hline
\textsc{Raven} & 1-shot & It's not the painting that sells for \$300 million dollars, it's the story behind the painting. \\
\hline
\textsc{Raven} & FiCL \tiny{[40-1]} & It's not the painting that sells for \$300 million dollars, it's the reputation of the artist. \\
\bottomrule
\toprule
\multicolumn{2}{l|}{\textbf{Question}} &  \textbf{When the society goes fully automatized, what will be the ``normal'' people needed for?} \\
\midrule
\multicolumn{2}{l|}{Human}  & Call centers... I am being totally serious. If your Sony TV is being completely fabricated by machines, shipped by machines, sold on amazon and delivered by machines, one of the few value adds companies will be able to deliver is GOOD customer service. So calling that 1-800 number and being instantly connected to a knowledgeable, friendly, human being who is going to patiently help you fix whatever issue you have (FYI... you forgot to plug it in). \\
\hline
\textsc{Atlas} & 1-shot & '$>$ ELI5: What will the society go fully automatized? $>$ \\
\hline
\textsc{Raven} & 1-shot & The ones who will program the robots.  \\
\hline
\textsc{Raven} & FiCL \tiny{[40-1]} & The ``normal'' people will be needed to program the robots. \\
\bottomrule
\toprule
\multicolumn{2}{l|}{\textbf{Question}} &  \textbf{Why is a 2 liter of Coca-cola \$1, a 12 oz bottle \$1.49 and the same size bottle of water \$1.99?} \\
\midrule
\multicolumn{2}{l|}{Human}  & Because people will pay. Plain and simple. Capitalism at its best. \\
\hline
\textsc{Atlas} & 1-shot & The price of the product is based on the cost of the packaging. \\
\hline
\textsc{Raven} & 1-shot & The cost of packaging is proportional to the size of the container. \\
\hline
\textsc{Raven} & FiCL \tiny{[40-1]} & The cost of packaging is the largest portion of the cost of a product. The larger the container, the less packaging is required. \\
\bottomrule
\toprule
\multicolumn{2}{l|}{\textbf{Question}} &  \textbf{If Matter Cannot Be Created or Destroyed... How do trees form?} \\
\midrule
\multicolumn{2}{l|}{Human}  & Say you have a box of Legos with 500 pieces. With those 500 pieces, you build a bulldozer. Then one day, you decide to rearrange those pieces from a bulldozer to a monster truck. It's still the same 500 pieces, just in a different layout. This is how trees are formed. The matter is just rearranged from other objects. Soil, nutrients, water, and even air. This is called ``Conservation of Mass.'' EDIT: added air. Thanks. :) \\
\hline
\textsc{Atlas} & 1-shot & Trees are made of carbon and oxygen. \\
\hline
\textsc{Raven} & 1-shot & Matter can be created or destroyed by combining two or more different types of matter. \\
\hline
\textsc{Raven} & FiCL \tiny{[40-1]} & The tree is a living organism. It is made up of matter. The tree is made up of carbon, hydrogen, oxygen, and water. \\
\bottomrule
\toprule
\multicolumn{2}{l|}{\textbf{Question}} &  \textbf{If the flu shot contains a non-active version of the flu, then why are ``flu like symptoms'' common for the days following?} \\
\midrule
\multicolumn{2}{l|}{Human}  & the non-active proteins trigger an immune system response which can involve some inflammatory system responses. once the immune system is activated for that infection your body will react more readily if the actual infection gets into you. \\
\hline
\textsc{Atlas} & 1-shot & The flu shot contains live virus particles. \\
\hline
\textsc{Raven} & 1-shot & The flu shot contains a live attenuated virus. This means that it is not able to infect the body, but it can still cause some side effects.  \\
\hline
\textsc{Raven} & FiCL \tiny{[40-1]} & The flu shot contains a weakened version of the virus. This weakened virus is able to trigger an immune response in the body. This immune response can cause flu like symptoms. \\
\bottomrule
\end{tabular}
\caption{Example outputs of \textsc{Atlas} and \textsc{Raven} 11B on long-form question answering.}
\label{table:eli5}
\end{table*}

\end{document}